\documentclass[runningheads]{llncs}
\usepackage{graphicx}

\usepackage{tikz}
\usepackage{comment}
\usepackage{amsmath,amssymb} 
\usepackage{color}

\newsavebox\CBox
\def\textBF#1{\sbox\CBox{#1}\resizebox{\wd\CBox}{\ht\CBox}{\textbf{#1}}}

\newcommand{\eg}{{\emph{e.g.}}}
\newcommand{\ie}{{\emph{i.e.}}}

\newcommand{\etc}{{\emph{etc}\ }}
\definecolor{grayhighlight}{RGB}{213,229,255}

\begin{document}
\pagestyle{headings}
\mainmatter
\def\ECCVSubNumber{100}  

\title{Learned Smartphone ISP on Mobile GPUs with Deep Learning, Mobile AI \& AIM 2022 Challenge: Report} 

\titlerunning{ECCV-22 submission ID \ECCVSubNumber}
\authorrunning{ECCV-22 submission ID \ECCVSubNumber}
\author{Anonymous ECCV submission}
\institute{Paper ID \ECCVSubNumber}

\author{Andrey Ignatov \and Radu Timofte \and
Shuai Liu \and Chaoyu Feng \and Furui Bai \and Xiaotao Wang \and Lei Lei \and
Ziyao Yi \and Yan Xiang \and Zibin Liu \and Shaoqing Li \and Keming Shi \and Dehui Kong \and Ke Xu \and
Minsu Kwon \and
Yaqi Wu \and Jiesi Zheng \and Zhihao Fan \and Xun Wu \and Feng Zhang \and
Albert No \and Minhyeok Cho \and
Zewen Chen \and Xiaze Zhang \and Ran Li \and Juan Wang \and Zhiming Wang \and
Marcos V. Conde \and Ui-Jin Choi \and
Georgy Perevozchikov \and Egor Ershov \and
Zheng Hui \and
Mengchuan Dong \and Xin Lou \and Wei Zhou \and Cong Pang \and
Haina Qin \and Mingxuan Cai
$^*$
}

\institute{}
\titlerunning{Learned Smartphone ISP on Mobile GPUs with Deep Learning}
\authorrunning{A. Ignatov, R. Timofte et al.}
\maketitle

\begin{abstract}

The role of mobile cameras increased dramatically over the past few years, leading to more and more research in automatic image quality enhancement and RAW photo processing. In this Mobile AI challenge, the target was to develop an efficient end-to-end AI-based image signal processing (ISP) pipeline replacing the standard mobile ISPs that can run on modern smartphone GPUs using TensorFlow Lite. The participants were provided with a large-scale Fujifilm UltraISP dataset consisting of thousands of paired photos captured with a normal mobile camera sensor and a professional 102MP medium-format FujiFilm GFX100 camera. The runtime of the resulting models was evaluated on the Snapdragon's 8 Gen 1 GPU that provides excellent acceleration results for the majority of common deep learning ops. The proposed solutions are compatible with all recent mobile GPUs, being able to process Full HD photos in less than 20-50 milliseconds while achieving high fidelity results. A detailed description of all models developed in this challenge is provided in this paper.

\end{abstract}

{\let\thefootnote\relax\footnotetext{%
$^*$Andrey Ignatov \textit{(andrey@vision.ee.ethz.ch)} and Radu Timofte \textit{(radu.timofte@uni-wuerzburg.de)} are the main Mobile AI \& AIM 2022 challenge organizers . The other authors participated in the challenge. \vspace{2mm} \\ Appendix \ref{sec:apd:team} contains the authors' team names and affiliations. \vspace{2mm} \\ Mobile AI 2022 Workshop website: \\ \url{https://ai-benchmark.com/workshops/mai/2022/}
}}

\section{Introduction}

\begin{figure*}[t!]
\centering
\setlength{\tabcolsep}{1pt}
\resizebox{\linewidth}{!}
{
\begin{tabular}{ccc}
\scriptsize{Visualized RAW Image}\normalsize & \scriptsize{MediaTek Dimensity 820 ISP Photo}\normalsize & \scriptsize{Fujifilm GFX 100 Photo}\normalsize\\
    \includegraphics[width=0.33\linewidth]{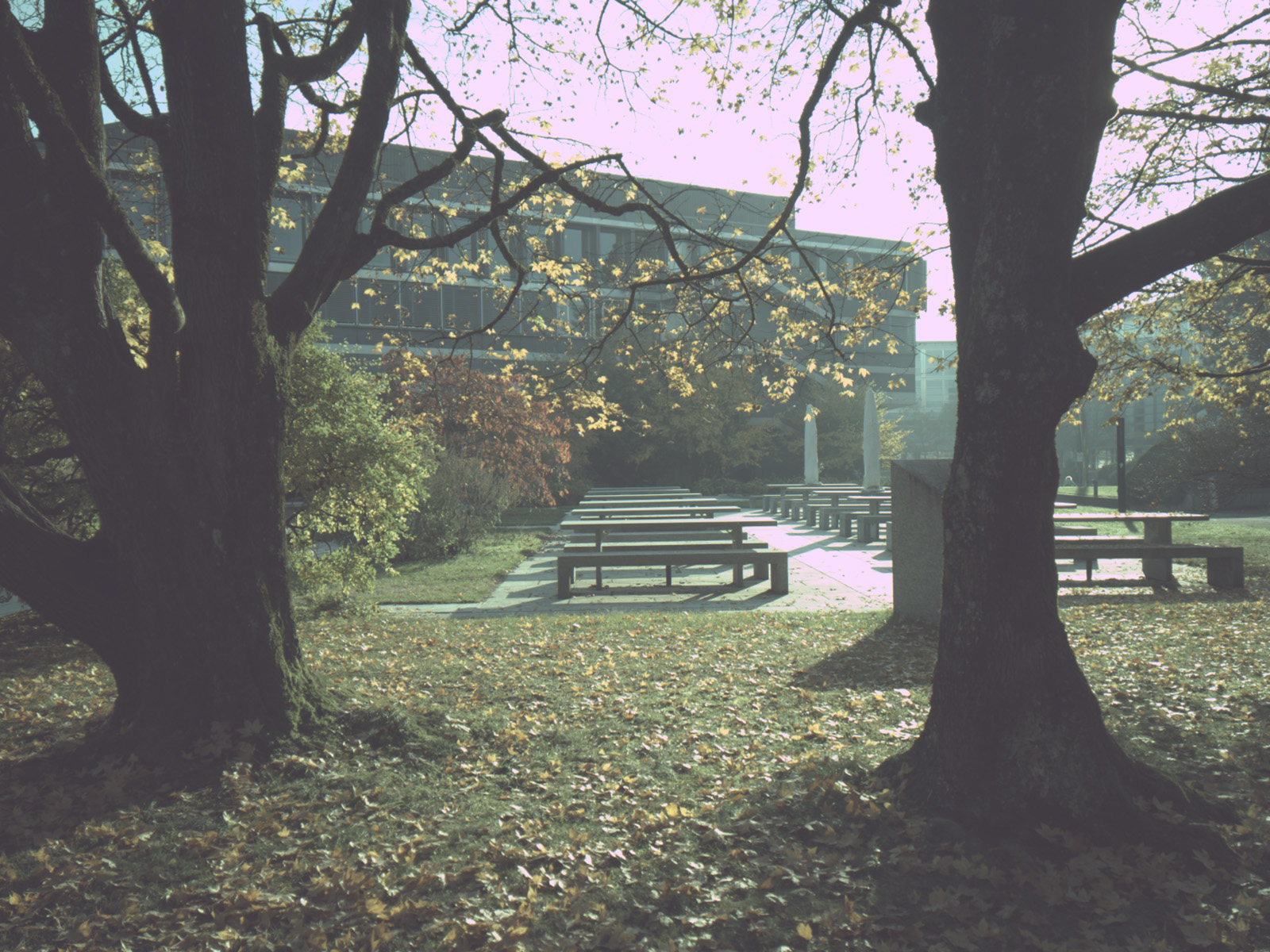}&
    \includegraphics[width=0.33\linewidth]{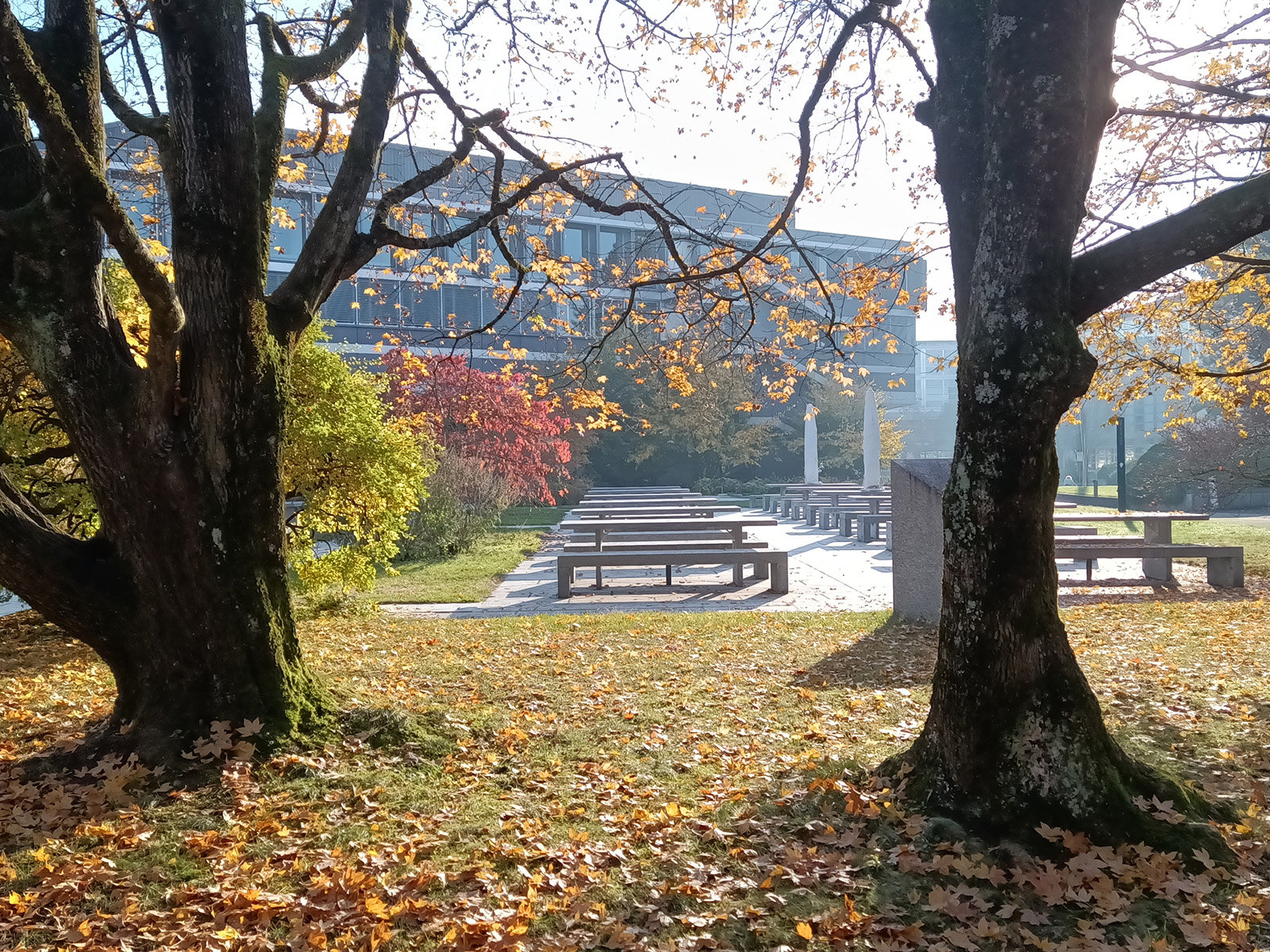}&
    \includegraphics[width=0.33\linewidth]{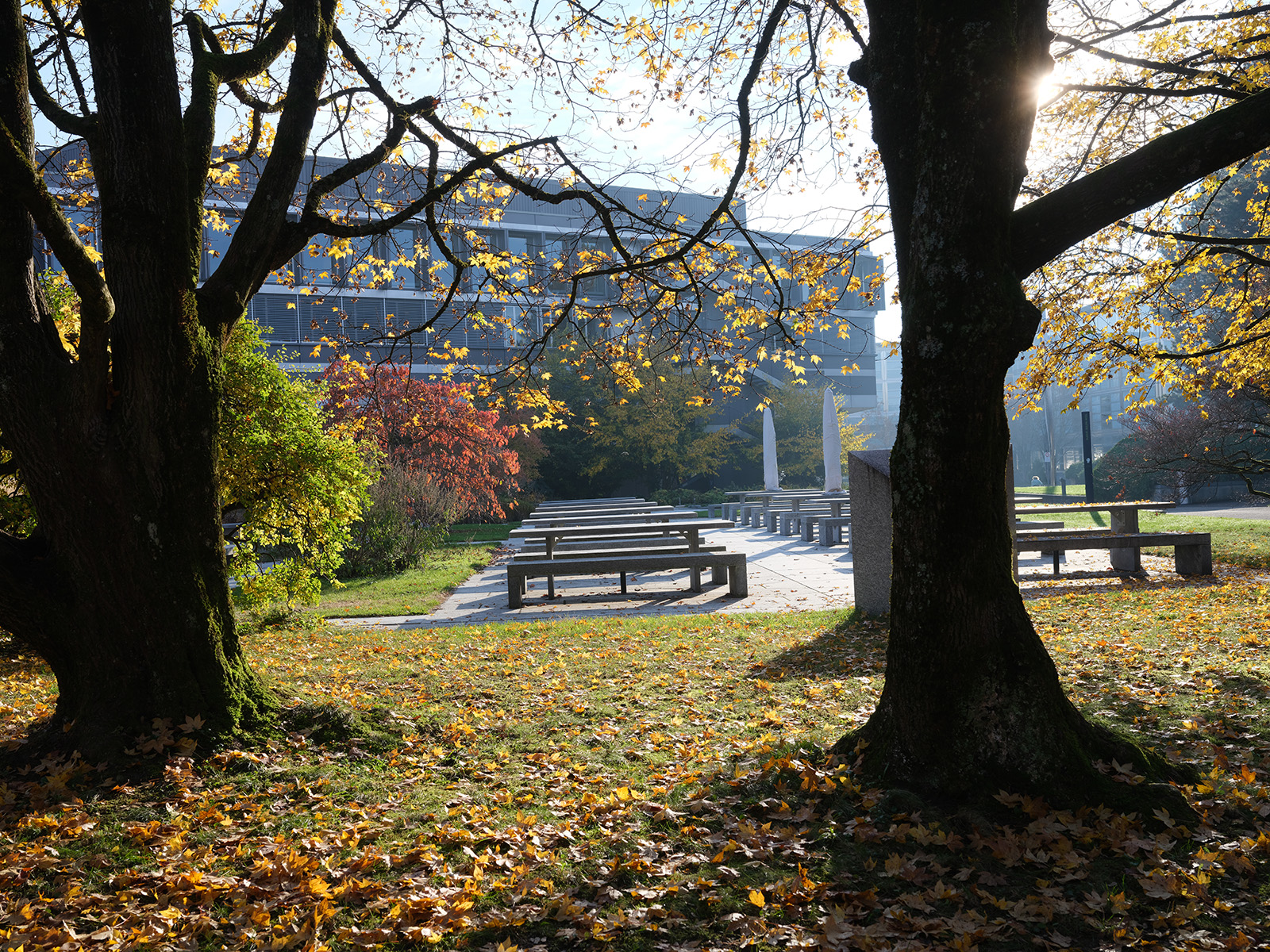} \\
    \includegraphics[width=0.33\linewidth]{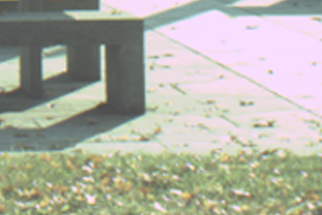}&
    \includegraphics[width=0.33\linewidth]{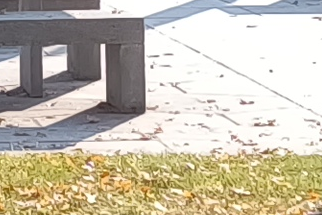}&
    \includegraphics[width=0.33\linewidth]{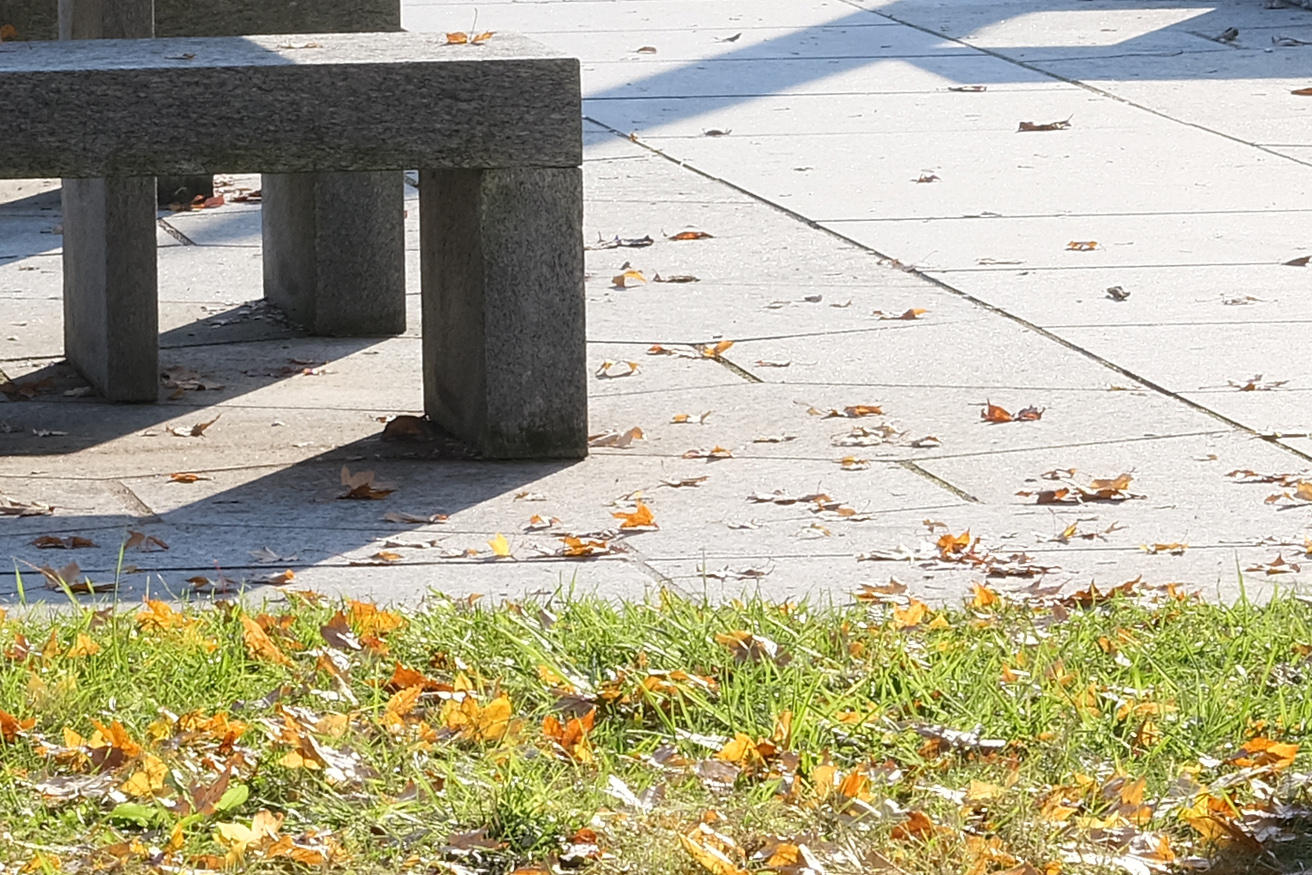} \\
\end{tabular}
}
\vspace{-0.2cm}
\caption{Example set of full-resolution images (top) and crops (bottom) from the collected Fujifilm UltraISP dataset. From left to right: original RAW visualized image, RGB image obtained with MediaTek's built-in ISP system, and Fujifilm GFX100 target photo.}
\vspace{-0.4cm}
\label{fig:example_photos}
\end{figure*}

Nowadays, the cameras are ubiquitous mainly due to the tremendous success and adoption of modern smartphones. Over the years, the quality of the smartphone cameras continuously improved due to advances in both hardware and software. Currently, due to their versatility, the critical improvements are coming from the advanced image processing algorithms employed, \eg, to perform color reconstruction or adjustment, noise removal, super-resolution, high dynamic range processing. The image enhancement task can be effectively solved with deep learning-based approaches. The critical part is the acquisition of appropriate (paired) low and high-quality ground truth images for training. For the first time the end-to-end mobile photo quality enhancement problem was tackled in~\cite{ignatov2017dslr,ignatov2018wespe}. The authors proposed to directly map the images from a low-quality smartphone camera to the higher-quality images from a high-end DSLR camera. The introduced DPED dataset was later employed in many competitions~\cite{ignatov2018pirm,ignatov2019ntire} and works~\cite{vu2018fast,lugmayr2019unsupervised,de2018fast,hui2018perception,huang2018range,liu2018deep} that significantly advanced the research on this problem. The major shortcoming of the proposed methods is that they are working on the images produced by cameras' built-in ISPs and, thus, they are not using a significant part of the original sensor data lost in the ISP pipeline. In~\cite{ignatov2020replacing} the authors proposed to replace the smartphone ISP with a deep neural network learned to map directly the RAW Bayer sensor data to the higher-quality images captured by a DSLR camera. For this, a \textit{Zurich RAW to RGB} dataset containing RAW-RGB image pairs from a mobile camera sensor and a high-end DSLR camera was collected. The proposed learned ISP reached the quality level of commercial ISP system of the Huawei P20 camera phone, and these results were further improved in~\cite{ignatov2020aim,dai2020awnet,silva2020deep,kim2020pynet,ignatov2019aim}. In this challenge, we use a more advanced FujiFlim UltraISP dataset~\cite{ignatov2022microisp,ignatov2022pynetv2} and additional efficiency-related constraints on the developed solutions. We target deep learning solutions capable to run on mobile GPUs. This is the second installment after the challenge conducted in conjunction with the Mobile AI 2021 CVPR workshop~\cite{ignatov2021learned}.

The deployment of AI-based solutions on portable devices usually requires an efficient model design based on a good understanding of the mobile processing units (\eg CPUs, NPUs, GPUs, DSP) and their hardware particularities, including their memory constraints. We refer to~\cite{ignatov2019ai,ignatov2018ai} for an extensive overview of mobile AI acceleration hardware, its particularities and performance. As shown in these works, the latest generations of mobile NPUs are reaching the performance of older-generation mid-range desktop GPUs. Nevertheless, a straightforward deployment of neural networks-based solutions on mobile devices is impeded by (i) a limited memory (\ie, restricted amount of RAM) and
(ii) a limited or lacking support of many common deep learning operators and layers. These impeding factors make the processing of high resolution inputs impossible with the standard NN models and require a careful adaptation or re-design to the constraints of mobile AI hardware. Such optimizations can employ a combination of various model techniques such as 16-bit / 8-bit~\cite{chiang2020deploying,jain2019trained,jacob2018quantization,yang2019quantization} and low-bit~\cite{cai2020zeroq,uhlich2019mixed,ignatov2020controlling,liu2018bi} quantization, network pruning and compression~\cite{chiang2020deploying,ignatov2020rendering,li2019learning,liu2019metapruning,obukhov2020t}, device- or NPU-specific adaptations, platform-aware neural architecture search~\cite{howard2019searching,tan2019mnasnet,wu2019fbnet,wan2020fbnetv2}, \etc.

The majority of competitions aimed at efficient deep learning models use standard desktop hardware for evaluating the solutions, thus the obtained models rarely show acceptable results when running on real mobile hardware with many specific constraints. In this \textit{Mobile AI challenge}, we take a radically different approach and propose the participants to develop and evaluate their models directly on mobile devices. The goal of this competition is to design a fast and performant deep learning-based solution for the learned smartphone ISP problem. For this, the participants were provided with the Fujifilm UltraISP dataset consisting of thousands of paired photos captured with a normal mobile camera sensor and a professional 102MP medium-format FujiFilm GFX100 camera. The efficiency of the proposed solutions was evaluated on the Snapdragon 8 Gen 1 mobile platform capable of accelerating floating-point and quantized neural networks. All solutions developed in this challenge are fully compatible with the TensorFlow Lite framework~\cite{TensorFlowLite2021}, thus can be efficiently executed on various Linux and Android-based IoT platforms, smartphones and edge devices.

\smallskip

This challenge is a part of the \textit{Mobile AI \& AIM 2022 Workshops and Challenges} consisting of the following competitions:

\small

\begin{itemize}
\item Learned Smartphone ISP on Mobile GPUs
\item Power Efficient Video Super-Resolution on Mobile NPUs~\cite{ignatov2022maivideosr}
\item Quantized Image Super-Resolution on Mobile NPUs~\cite{ignatov2022maisuperres}
\item Efficient Single-Image Depth Estimation on Mobile Devices~\cite{ignatov2022maidepth}
\item Realistic Bokeh Effect Rendering on Mobile GPUs~\cite{ignatov2022maibokeh}
\item Super-Resolution of Compressed Image and Video~\cite{yang2022aim}
\item Reversed Image Signal Processing and RAW Reconstruction~\cite{conde2022aim}
\item Instagram Filter Removal~\cite{kinli2022aim}
\end{itemize}

\noindent The results and solutions obtained in the previous \textit{MAI 2021 Challenges} are described in our last year papers:

\small

\begin{itemize}
\item Learned Smartphone ISP on Mobile NPUs~\cite{ignatov2021learned}
\item Real Image Denoising on Mobile GPUs~\cite{ignatov2021fastDenoising}
\item Quantized Image Super-Resolution on Mobile NPUs~\cite{ignatov2021real}
\item Real-Time Video Super-Resolution on Mobile GPUs~\cite{romero2021real}
\item Single-Image Depth Estimation on Mobile Devices~\cite{ignatov2021fastDepth}
\item Quantized Camera Scene Detection on Smartphones~\cite{ignatov2021fastSceneDetection}
\end{itemize}

\normalsize


\begin{figure*}[t!]
\centering
\setlength{\tabcolsep}{1pt}
\resizebox{0.96\linewidth}{!}
{
\includegraphics[width=1.0\linewidth]{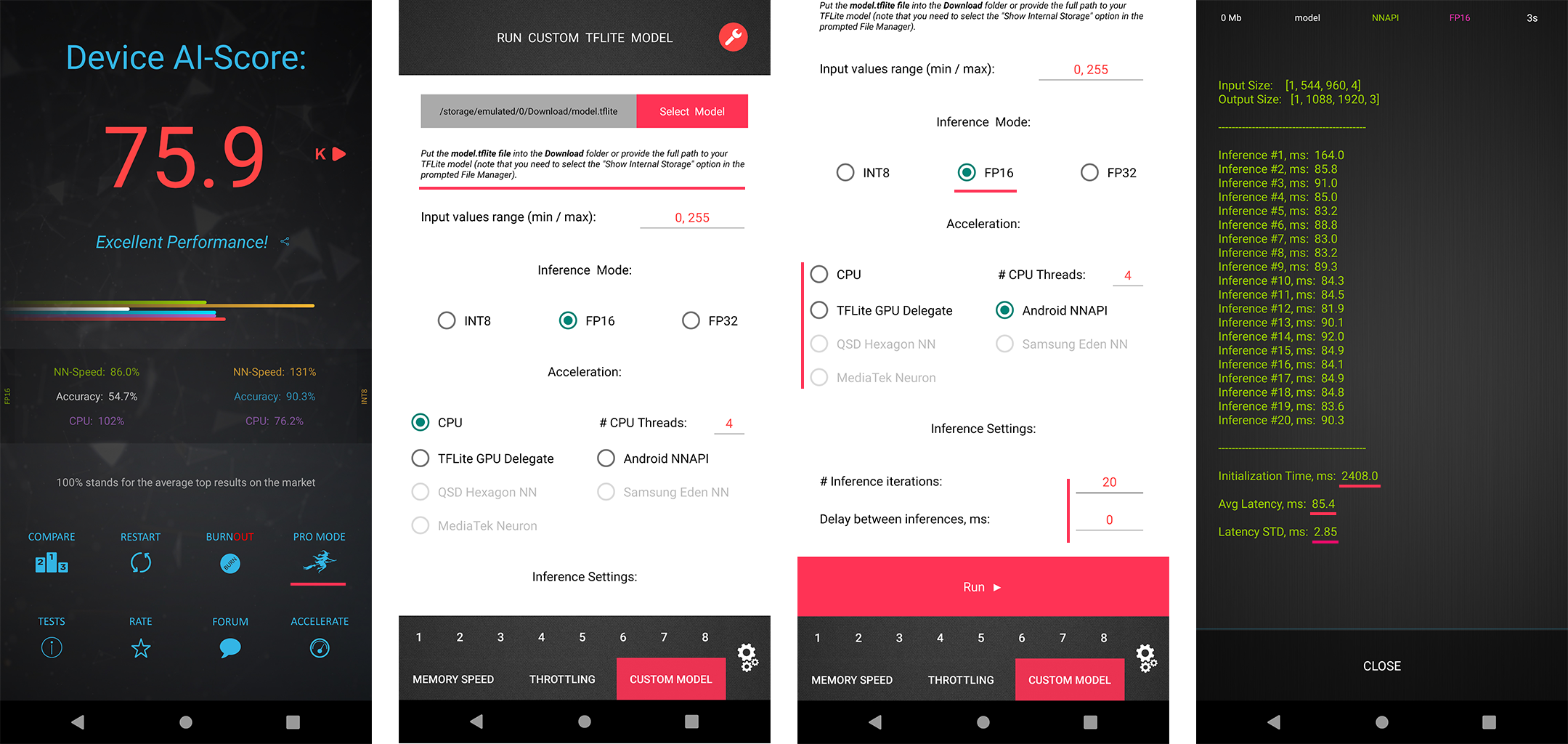}
}
\vspace{0.2cm}
\caption{Loading and running custom TensorFlow Lite models with AI Benchmark application. The currently supported acceleration options include Android NNAPI, TFLite GPU, Hexagon NN, Qualcomm QNN, MediaTek Neuron and Samsung ENN delegates as well as CPU inference through TFLite or XNNPACK backends. The latest app version can be downloaded at \url{https://ai-benchmark.com/download}.}
\label{fig:ai_benchmark_custom}
\end{figure*}

\section{Challenge}

In order to design an efficient and practical deep learning-based solution for the considered task that runs fast on mobile devices, one needs the following tools:

\begin{enumerate}
\item A large-scale high-quality dataset for training and evaluating the models. Real, not synthetically generated data should be used to ensure a high quality of the obtained model;
\item An efficient way to check the runtime and debug the model locally without any constraints as well as the ability to check the runtime on the target evaluation platform.
\end{enumerate}

This challenge addresses all the above issues. Real training data, tools, and runtime evaluation options provided to the challenge participants are described in the next sections.

\subsection{Dataset}

In this challenge, we use the Fujifilm UltraISP dataset collected using the Fujifilm GFX100 medium format 102 MP camera capturing the target high-quality images, and a popular Sony IMX586 Quad Bayer mobile camera sensor that can be found in tens of mid-range and high-end mobile devices released in the past 3 years. The Sony sensor was mounted on the MediaTek Dimensity 820 development board, and was capturing both raw and processed (by its built-in ISP system) 12MP images. The Dimensity board was rigidly attached to the Fujifilm camera, and they were shooting photos synchronously to ensure that the image content is identical. The dataset contains over 6 thousand daytime image pairs captured at a wide variety of places with different illumination and weather conditions. An example set of full-resolution photos from the Fujifilm UltraISP dataset is shown in Fig.~\ref{fig:example_photos}.
As the collected RAW-RGB image pairs were not perfectly aligned, they were initially matched using the state-of-the-art deep learning based dense matching algorithm~\cite{truong2021learning} to extract 256$\times$256 pixel patches from the original photos. It should be mentioned that all alignment operations were performed on Fujifilm RGB images only, therefore RAW photos from the Sony sensor remained unmodified, exhibiting the same values as read from the sensor.

\subsection{Local Runtime Evaluation}

When developing AI solutions for mobile devices, it is vital to be able to test the designed models and debug all emerging issues locally on available devices. For this, the participants were provided with the \textit{AI Benchmark} application~\cite{ignatov2018ai,ignatov2019ai} that allows to load any custom TensorFlow Lite model and run it on any Android device with all supported acceleration options. This tool contains the latest versions of \textit{Android NNAPI, TFLite GPU, Hexagon NN, Qualcomm QNN, MediaTek Neuron} and \textit{Samsung ENN} delegates, therefore supporting all current mobile platforms and providing the users with the ability to execute neural networks on smartphone NPUs, APUs, DSPs, GPUs and CPUs.

\smallskip

To load and run a custom TensorFlow Lite model, one needs to follow the next steps:

\begin{enumerate}
\setlength\itemsep{0mm}
\item Download AI Benchmark from the official website\footnote{\url{https://ai-benchmark.com/download}} or from the Google Play\footnote{\url{https://play.google.com/store/apps/details?id=org.benchmark.demo}} and run its standard tests.
\item After the end of the tests, enter the \textit{PRO Mode} and select the \textit{Custom Model} tab there.
\item Rename the exported TFLite model to \textit{model.tflite} and put it into the \textit{Download} folder of the device.
\item Select mode type \textit{(INT8, FP16, or FP32)}, the desired acceleration/inference options and run the model.
\end{enumerate}

\noindent These steps are also illustrated in Fig.~\ref{fig:ai_benchmark_custom}.

\subsection{Runtime Evaluation on the Target Platform}

In this challenge, we use the the \textit{Qualcomm Snapdragon 8 Gen 1} mobile SoC as our target runtime evaluation platform. The considered chipset demonstrates very decent AI Benchmark scores and can be found in the majority of flagship Android smartphones released in 2022. It can efficiently accelerate floating-point networks on its Adreno 730 GPU with a theoretical FP16 performance of 5 TFLOPS. The models were parsed and accelerated using the TensorFlow Lite GPU delegate~\cite{lee2019device} demonstrating the best performance on this platform when using general deep learning models. All final solutions were tested using the aforementioned AI Benchmark application.

\subsection{Challenge Phases}

\begin{table*}[t!]
\centering
\resizebox{\linewidth}{!}
{
\begin{tabular}{l|c|cc|cc|cc|c}
\hline
Team \, & \, Author \, & \, Framework \, & \, Model Size, MB \, & \, PSNR$\uparrow$ \, & \, SSIM$\uparrow$ & \, CPU Runtime, ms $\downarrow$ \, & GPU Runtime, ms $\downarrow$ \, & \, Final Score \\
\hline
\hline

MiAlgo & mialgo\_ls & TensorFlow & 0.014 & 23.33 & 0.8516 & \textBF{135} & \textBF{6.8} & \textBF{14.87} \\
ENERZAi	& MinsuKwon & TensorFlow & 0.077 &  23.8 & 0.8652 & 208 & 18.9 & 10.27 \\
HITZST01 & Jaszheng & Keras / TensorFlow & 0.060 &  23.89 & 0.8666 & 712 & 34.3 & 6.41 \\
MINCHO	&	Minhyeok & TensorFlow & 0.067 &  23.65 & 0.8658 & 886 & 41.5 & 3.8 \\
ENERZAi	&	MinsuKwon & TensorFlow & 4.5 &  24.08 & 0.8778 & 45956 & 212 & 1.35 \\
HITZST01	&	Jaszheng & Keras / TensorFlow & 1.2 &  \textBF{24.09} & 0.8667 & 4694 & 482 & 0.6 \\
JMU-CVLab	&	nanashi & Keras / TensorFlow & 0.041 &  23.22 & 0.8281 & 3487 & 182 & 0.48 \\
rainbow	&	zheng222 & TensorFlow & 1.0 &  21.66 & 0.8399 & 277 & 28 & 0.36 \\
CASIA 1st	&	Zevin & PyTorch / TensorFlow & 205 &  \textBF{24.09} & \textBF{0.884} & 14792 & 1044 & 0.28 \\
MiAlgo	&	mialgo\_ls & \, PyTorch / TensorFlow \, & 117 &  23.65 & 0.8673 & 15448 & 1164 & 0.14 \\
DANN-ISP	&	gosha20777 & TensorFlow & 29.4 &  23.1 & 0.8648 & 97333 & 583 & 0.13 \\
\hline
Multimedia $^*$	&	lillythecutie & PyTorch/ OpenVINO & 0.029 &  23.96 & 0.8543 & 293 & 11.4 & 21.24 \\
\hline
SKD-VSP	& \, dongdongdong \, & PyTorch / TensorFlow & 78.9 &  24.08 & 0.8778 & $>$ 10 min & Failed & N.A. \\
CHannel Team \,	&	sawyerk2212 & PyTorch / TensorFlow & 102.0 &  22.28 & 0.8482 & $>$ 10 min & Failed & N.A. \\
\hline
\hline
MicroISP 1.0~\cite{ignatov2022microisp} & \, Baseline \, & TensorFlow & 0.152 &  23.87 & 0.8530 & 973 & 23.1 & 9.25 \\
MicroISP 0.5~\cite{ignatov2022microisp} & \, Baseline \, & TensorFlow & 0.077 &  23.60 & 0.8460 & 503 & 15.6 & 9.43 \\
PyNET-V2 Mobile~\cite{ignatov2022pynetv2} \,	& Baseline & TensorFlow & 3.6 &  24.72 & 0.8783 & 8342 & 194 & 3.58 \\
\end{tabular}
}
\vspace{2.6mm}
\caption{\small{Mobile AI 2022 learned smartphone ISP challenge results and final rankings. The runtime values were obtained on Full HD (1920$\times$1088) resolution images on the Snapdragon 8 Gen 1 mobile platform. The results of the MicroISP and PyNET-V2 Mobile models are provided for the reference. $^*$~The solution submitted by team \textit{Multimedia} had corrupted weights due to incorrect model conversion, this issue was fixed after the end of the challenge.}}
\label{tab:results}
\end{table*}

The challenge consisted of the following phases:

\vspace{-0.8mm}
\begin{enumerate}
\item[I.] \textit{Development:} the participants get access to the data and AI Benchmark app, and are able to train the models and evaluate their runtime locally;
\item[II.] \textit{Validation:} the participants can upload their models to the remote server to check the fidelity scores on the validation dataset, and to compare their results on the validation leaderboard;
\item[III.] \textit{Testing:} the participants submit their final results, codes, TensorFlow Lite models, and factsheets.
\end{enumerate}
\vspace{-0.8mm}

\subsection{Scoring System}

All solutions were evaluated using the following metrics:

\vspace{-0.8mm}
\begin{itemize}
\setlength\itemsep{-0.2mm}
\item Peak Signal-to-Noise Ratio (PSNR) measuring fidelity score,
\item Structural Similarity Index Measure (SSIM), a proxy for perceptual score,
\item The runtime on the target Snapdragon 8 Gen 1 platform.
\end{itemize}
\vspace{-0.8mm}

In this challenge, the participants were able to submit their final models to two tracks. In the first track, the score of each final submission was evaluated based on the next formula ($C$ is a constant normalization factor):

\smallskip
\begin{equation*}
\text{Final Score} \,=\, \frac{2^{2 \cdot \text{PSNR}}}{C \cdot \text{runtime}},
\end{equation*}
\smallskip

In the second track, all submissions were evaluated only based on their visual results as measured by the corresponding Mean Opinion Scores (MOS). This was done to allow the participants to develop larger and more powerful models for the considered task.

During the final challenge phase, the participants did not have access to the test dataset. Instead, they had to submit their final TensorFlow Lite models that were subsequently used by the challenge organizers to check both the runtime and the fidelity results of each submission under identical conditions. This approach solved all the issues related to model overfitting, reproducibility of the results, and consistency of the obtained runtime/accuracy values.

\begin{table*}[t!]
\centering
\resizebox{\linewidth}{!}
{
\begin{tabular}{l|c|cc|cc|c}
\hline
Team \, & \, Author \, & \, Framework \, & \, Model Size, MB \, & \, PSNR$\uparrow$ \, & \, SSIM$\uparrow$ & \, MOS Score \\
\hline
\hline

HITZST01	&	Jaszheng & Keras / TensorFlow & 1.2 &  \textBF{24.09} & 0.8667 & 3.1 \\
ENERZAi	&	MinsuKwon & TensorFlow & 4.5 &  24.08 & 0.8778 & 3.1 \\
CASIA 1st	&	Zevin & PyTorch / TensorFlow & 205 &  \textBF{24.09} & \textBF{0.884} & 3.0 \\
Multimedia	& lillythecutie & PyTorch/ OpenVINO & 0.029 &  23.96 & 0.8543 & 3.0 \\
HITZST01 & Jaszheng & Keras / TensorFlow & 0.060 &  23.89 & 0.8666 & 3.0 \\
MINCHO	&	Minhyeok & TensorFlow & 0.067 &  23.65 & 0.8658 & 3.0 \\
ENERZAi	& \, MinsuKwon \, & TensorFlow & 0.077 &  23.8 & 0.8652 & 2.8 \\
DANN-ISP	&	gosha20777 & TensorFlow & 29.4 &  23.1 & 0.8648 & 2.8 \\
MiAlgo & mialgo\_ls & TensorFlow & 0.014 & 23.33 & 0.8516 & 2.5 \\
JMU-CVLab \,	&	nanashi & Keras / TensorFlow & 0.041 &  23.22 & 0.8281 & 2.3 \\
MiAlgo	&	mialgo\_ls & \, PyTorch / TensorFlow \, & 117 &  23.65 & 0.8673 & 2.2 \\
\end{tabular}
}
\vspace{2.6mm}
\caption{\small{Mean Opinion Scores (MOS) of all solutions submitted during the final phase of the MAI 2022 challenge and achieving a PSNR score of at least 23 dB. Visual results were assessed based on the reconstructed 12MP full resolution images.}}
\label{tab:results_mos}
\end{table*}

\section{Challenge Results}

From the above 140 registered participants, 11 teams entered the final phase and submitted valid results, TFLite models, codes, executables, and factsheets. The proposed methods are described in Section~\ref{sec:solutions}, and the team members and affiliations are listed in Appendix~\ref{sec:apd:team}.

\subsection{Results and Discussion}

Tables~\ref{tab:results} and~\ref{tab:results_mos} demonstrate the fidelity, runtime and MOS results of all solutions submitted during the final test phase. Models submitted to the 1st and 2nd challenge tracks were evaluated together since the participants had to upload the corresponding TensorFlow Lite models in both cases. In the 1st tack, the overall best results were achieved by team \textit{Multimedia}. The authors proposed a novel \textit{eReopConv} layer that consists of a large number of convolutions that are fused during the final model exporting stage to improve the its runtime while maintaining the fidelity scores. This approach turned out to be very efficient as the model submitted by this team was able to achieve one of the best PSNR and MOS scores as well as a runtime of less than 12 ms on the target Snapdragon platform. Unfortunately, the original TFLite file submitted by this team had corrupted weights caused by incorrect model conversion (the initial PyTorch model was converted to TFLite via ONNX), and this issue was fixed only after the end of the challenge.

The best runtime on the Snapdragon 8 Gen 1 was achieved by team \textit{MiAlgo}, which solution is able to process one Full HD resolution photo on its GPU under 7 milliseconds. This efficiency was achieved due to a very shallow structure of the proposed 3-layer neural network, which architecture was inspired by the last year MAI challenge winner~\cite{ignatov2021learned}. The visual results obtained by this model are also satisfactory, though significantly fall behind the results of the solution from team \textit{Multimedia}. The second best result in the 1st challenge track was obtained by team \textit{ENERZAi} that proposed a UNet-based model with channel attention blocks. The final structure of this solution was obtained using the neural architecture search modified to take the computational complexity of the model as an additional key penalty parameter.

After evaluating the visual quality of the proposed models, we obtained quite similar MOS scores for half of the submissions. A detailed inspection of the results revealed that their overall quality can be generally considered as relatively comparable, though none of the models was able to perform an ideal image reconstruction: each model had some issues either with color rendition or with noise suppression / texture rendering. Solutions with a MOS score of less than 2.8 were usually having several issues or were exhibiting noticeable image corruptions. These results highlighted again the difficulty of an accurate assessment of the results obtained in learned ISP task as the conventional fidelity metrics are often not indicating the real image quality.

\section{Challenge Methods}
\label{sec:solutions}

\noindent This section describes solutions submitted by all teams participating in the final stage of the MAI 2022 Learned Smartphone ISP challenge.

\subsection{MiAlgo}

\begin{figure}[h!]
\centering
\resizebox{1.0\linewidth}{!}
{
\includegraphics[width=0.45\linewidth]{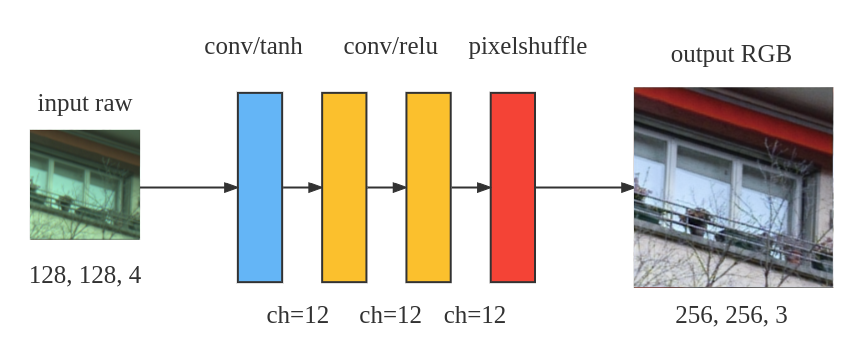} \vspace{2mm}
\includegraphics[width=0.45\linewidth]{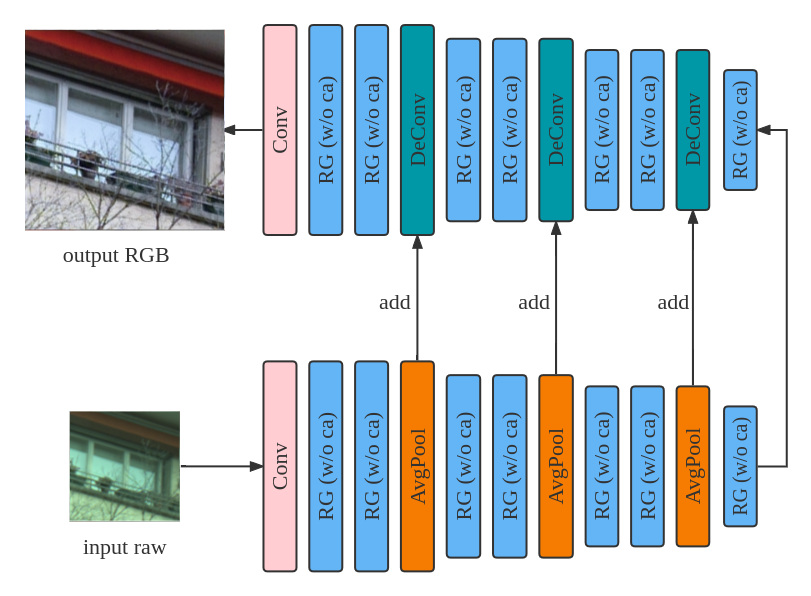}
}
\caption{\small{Model architectures proposed by team MiAlgo for the 1st (left) and 2nd (right) challenge tracks.}}
\label{fig:MiAlgo}
\end{figure}

For track 1, team MiAlgo proposed a smaller three-convolution structure based on last year's MAI 2021 winning solution~\cite{ignatov2021learned} (Fig. ~\ref{fig:MiAlgo}). The authors reduced the number of convolutional channels from 16 to 12, which also decreased the inference time by about 1/4. Besides that, the authors additionally used the distillation technique to remove the misalignment of some raw-RGB pairs in order to make the model converge better, which improved the PSNR score by about 0.3 dB. The model was trained for 10K epochs with L1 loss. The parameters were optimized with the Adam~\cite{kingma2014adam} algorithm using a batch size of 32 and a learning rate of 1e$-$4 that was decreased within the training.

For track 2, the authors proposed a 4-level UNet-based structure (Fig. ~\ref{fig:MiAlgo}, right). Several convolutional layers in the UNet~\cite{ronneberger2015u} architecture were replaced with a residual group (RG, without channel attention layer) from RCAN~\cite{zhang2018image} to enhance the reconstruction ability of the network. The authors used average pooling for the down-sampling layer and deconvolution for the up-sampling layer. The number of channels for each model level is 32, 64, 128, and 256 respectively. The model was first trained for 2K epochs with L1 loss, and then fine-tuned for about 2K epochs with the L1 and VGG loss functions. The initial learning rate was set to 1e--4 and was decreased within the training.

\subsection{Multimedia}

\begin{figure}[h!]
\centering
\includegraphics[width=0.9\linewidth]{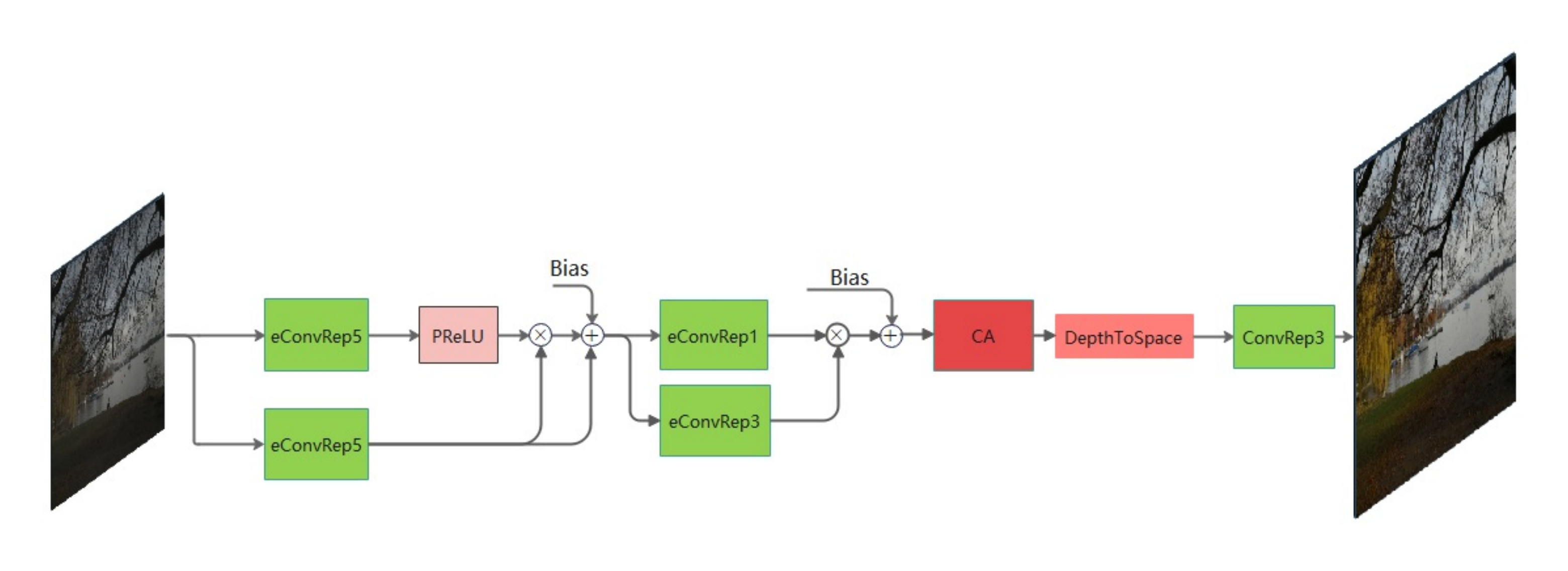} \vspace{4mm} \\
\includegraphics[width=0.8\linewidth]{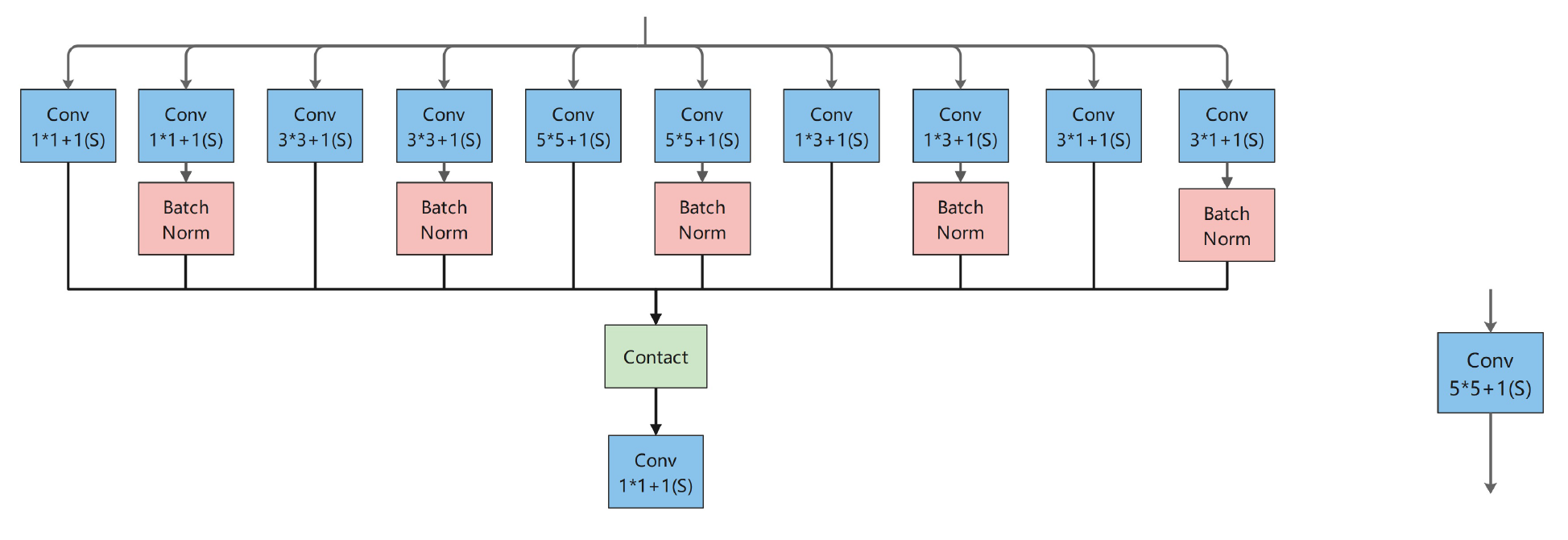} \vspace{6mm} \\
\includegraphics[width=0.6\linewidth]{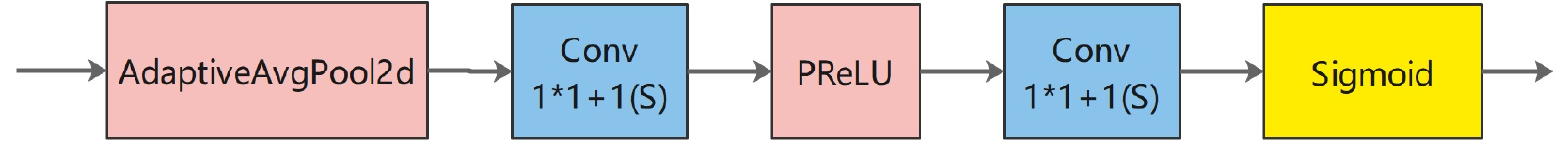}
\caption{\small{The overall model architecture (top), the structure of the \textit{eReopConv} block during the training and inference stages (middle), and the architecture of the CA module (bottom) proposed by team Multimedia.}}
\label{fig:Multimedia}
\end{figure}

Inspired by a series of research on model re-parameterization from Ding~\cite{ding2021repvgg,ding2021repmlp,ding2021resrep}, team Multimedia proposed an \textit{enormous Re-parameter Convolution (eReopConv)} layer to replace the standard convolution. eReopConv has a large structure during training to learn superior proprieties but transforms into a less structured block during inference and retains these proprieties. The training and inference structures are shown in Fig.~\ref{fig:Multimedia}: unlike the RepConv in the RepVGG~\cite{ding2021repvgg} that contain a $3\times3$ and $1\times1$ convolution layers in two branches during the training procedure, this method retains multi-branch convolution layers with different kernel sizes. \textit{E.g.,} eRepConv with $5\times5$ kernel size has ten convolution layers with a kernel size ranging from $1\times1$ to $5\times5$ to collect enough information from different receptive fields. During the inference procedure, the training parameters are re-parameterized by continuous linear transforms.

As many spatial features are extracted by the eRepConv, a spatial attention mechanism could effectively improve the network performance. However, classical spatial attention blocks like~\cite{woo2018cbam} would increase the computational complexity because of the max pooling and average pooling operations. To save the extra costs, the authors utilize another eRepconv to produce a fine-granularity spatial attention block that has a specific attention matrix for each spatial feature in different channels. A learnable parameter is also added as a bias to expand this nonlinear expressivity. Although the fine-granularity spatial attention improves the performance, too many multiple operators and all channels fused by one convolution operator will inevitably make the model hard to train. To fix this, channel attention is added at the end of the network to help the network to converge since it could identify the importance of each channel.

Besides the Charbonnier loss and the cosine similarity loss, the authors propose to use an additional patch loss to enhance the image quality by considering the patch level information. In the patch loss, the ground truth $y$ and the generated $y'$ images are divided into a set of patches ${\{y_{p1},y_{p2},...,y_{pn}\}}$ and $\{y'_{p1}, y'_{p2},...y'_{pn}\}$,  then the mean and the variance of the difference between the generated and the ground truth patches is calculated and used as a weight for each pixel in the patch. The exact formulation of this loss function is:
\begin{equation}
\label{loss}
    L_{P} = \sum^{n}_{i=0}e^{mean(y_{pi}-y'_{pi})^{-1} +var(y_{pi}-y'_{pi})^-1} \times|y-y'|,
\end{equation}
where \textit{mean} represents how big the difference between the two images is, and \textit{var} reflects whether the two images have similar changing rates.

The model is trained in two stages. During the warming-up training stage, besides the RAW-to-RGB transformation task, the model also learns to perform masked raw data recovery: raw image is divided into a set of $3\times3$ patches, 50\% of them are randomly masked out, and the network is trained to recover the masked patches. This MAE-liked~\cite{he2022masked} strategy has two benefits. First, defective pixels often occur in the raw data, and it is quite useful to be able to recover these pixels. Secondly, since it is harder for a network to perform these two tasks simultaneously, the resulting model is usually more robust. The learning rate of the warming-up stage is set to 1e--6, and the model is trained with the L1 loss only. Next, in the second normal train stage the learning rate is set to 1e--3 and decayed with a cosine annealing scheduler, and the model is optimized with the Adam for only 350 epochs.

\subsection{ENERZAi}

\begin{figure}[h!]
\centering
\includegraphics[width=1.0\linewidth]{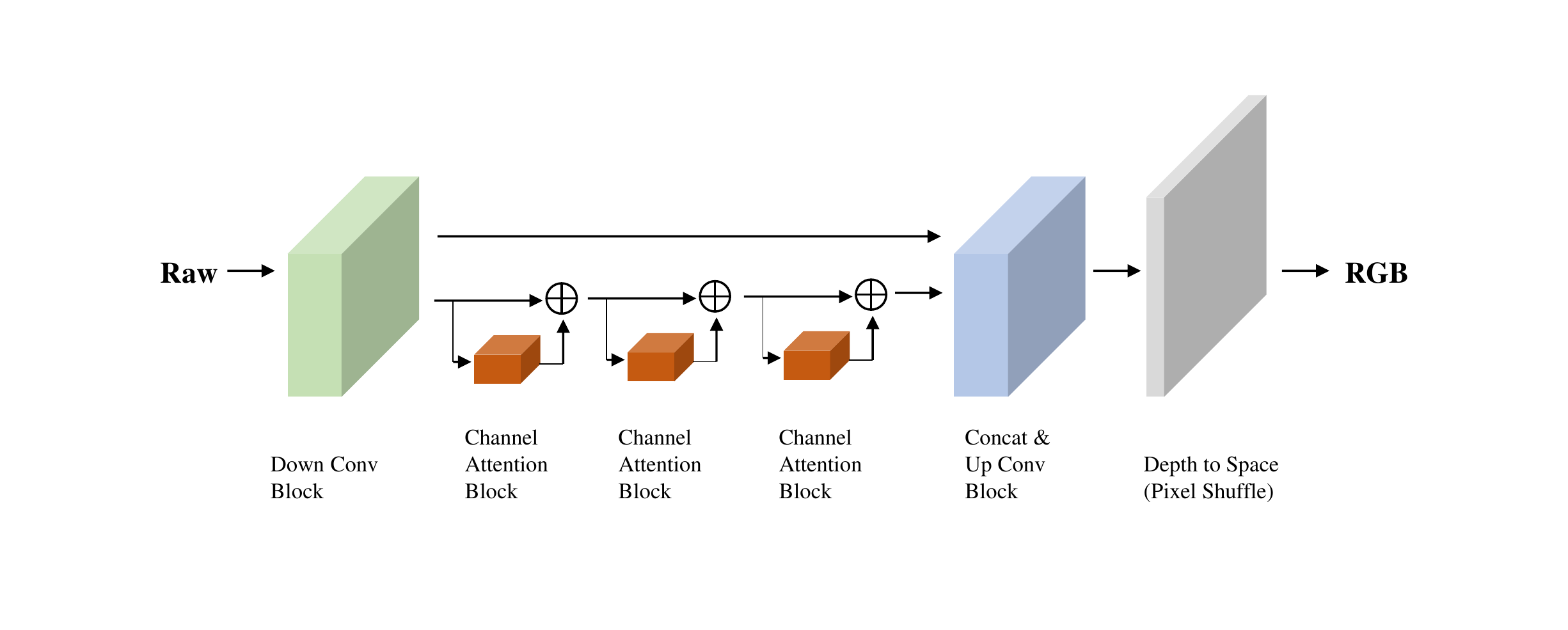}
\includegraphics[width=1.0\linewidth]{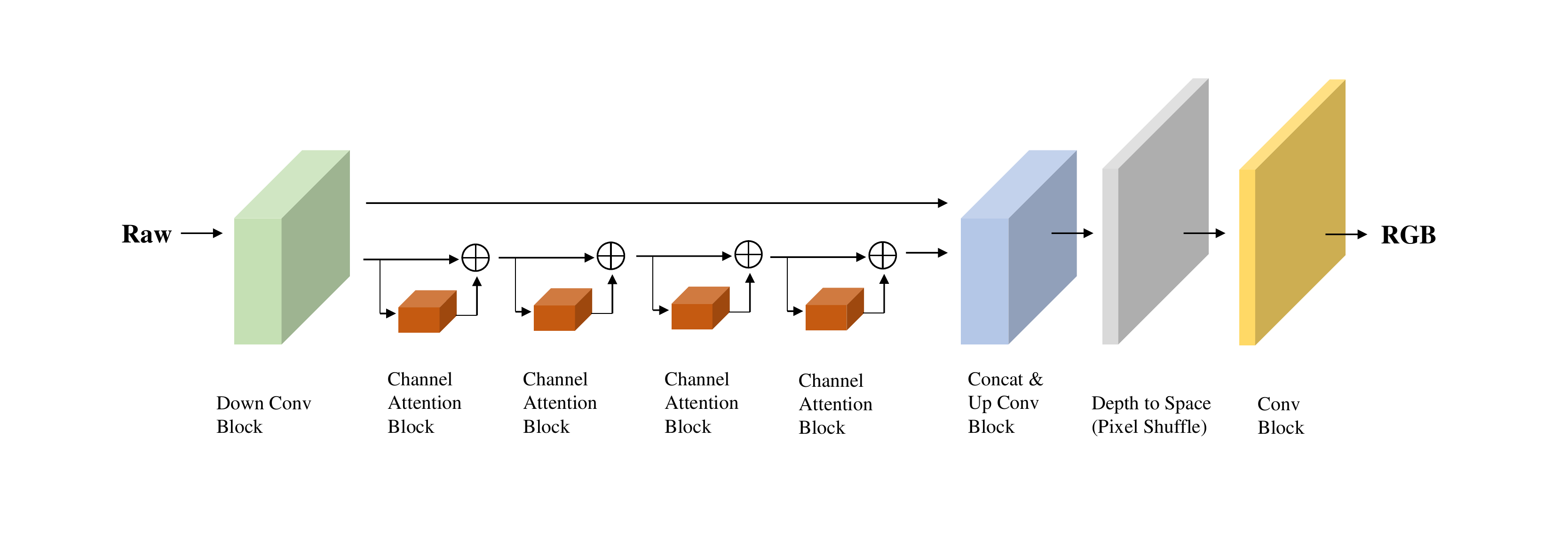}
\caption{\small{Model designs obtained by team ENERZAi for the 1st (top) and 2nd (bottom) challenge tracks using the neural architecture search.}}
\label{fig:ENERZAi}
\end{figure}

For track 1, team ENERZAi first constructed a search space for the target model architecture based on the UNet~\cite{ronneberger2015u} and Hourglass architectures. Since a right balance between image colors is important when constructing RGB images from raw data, a channel attention module~\cite{zhang2018image} was also included in the model space. Optimal model design (Fig.~\ref{fig:ENERZAi}) was obtained using an architecture search algorithm that was based on the evolutionary algorithm~\cite{real2019regularized} and was taking latency into account by penalizing computationally heavy models.

The authors used L1 loss, MS-SIM and ResNet50-based perceptual loss functions to train the model. In addition, the authors developed a differentiable approximate histogram feature inspired by the Histogan~\cite{afifi2021histogan} paper to consider the color distribution of the constructed RGB image. For each R, G, and B channel, the histogram differences between the constructed image and the target image were calculated and added to the total loss. The Adam optimizer with a learning rate of $0.001$ was used for training the model, and the learning rate was halved each several epochs.

For track 2 (Fig.~\ref{fig:ENERZAi}, bottom), the authors used the same approach but without penalizing computationally heavy models, which resulted in the increased number of convolutional layers and channel sizes.

\subsection{HITZST01}

\begin{figure}[h!]
\centering
\includegraphics[width=0.9\linewidth]{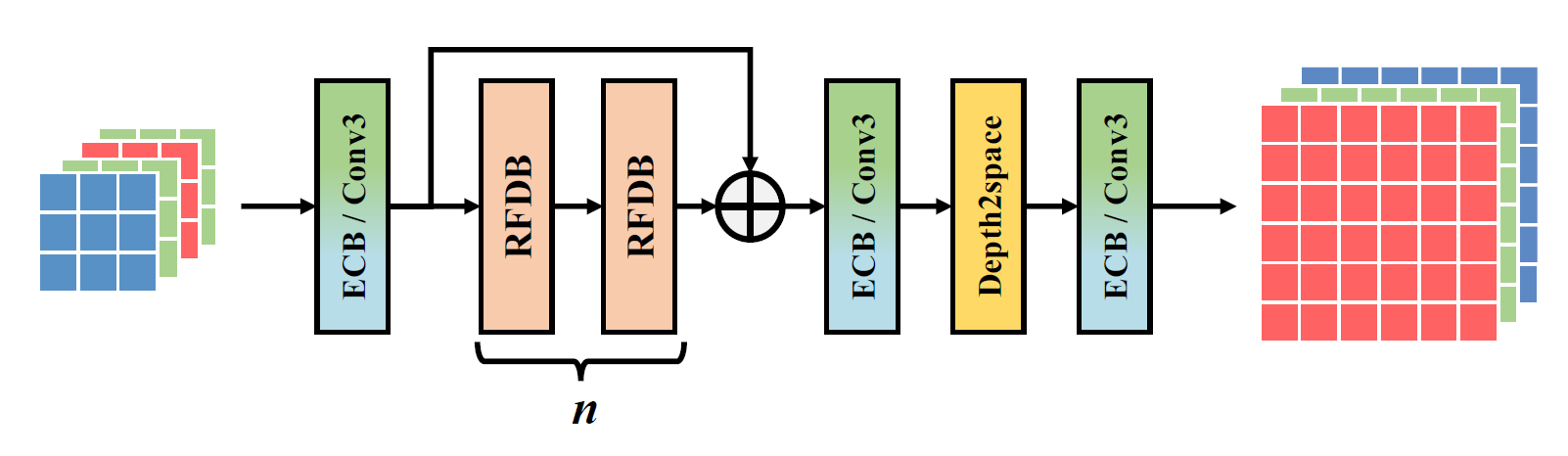}
\includegraphics[width=0.7\linewidth]{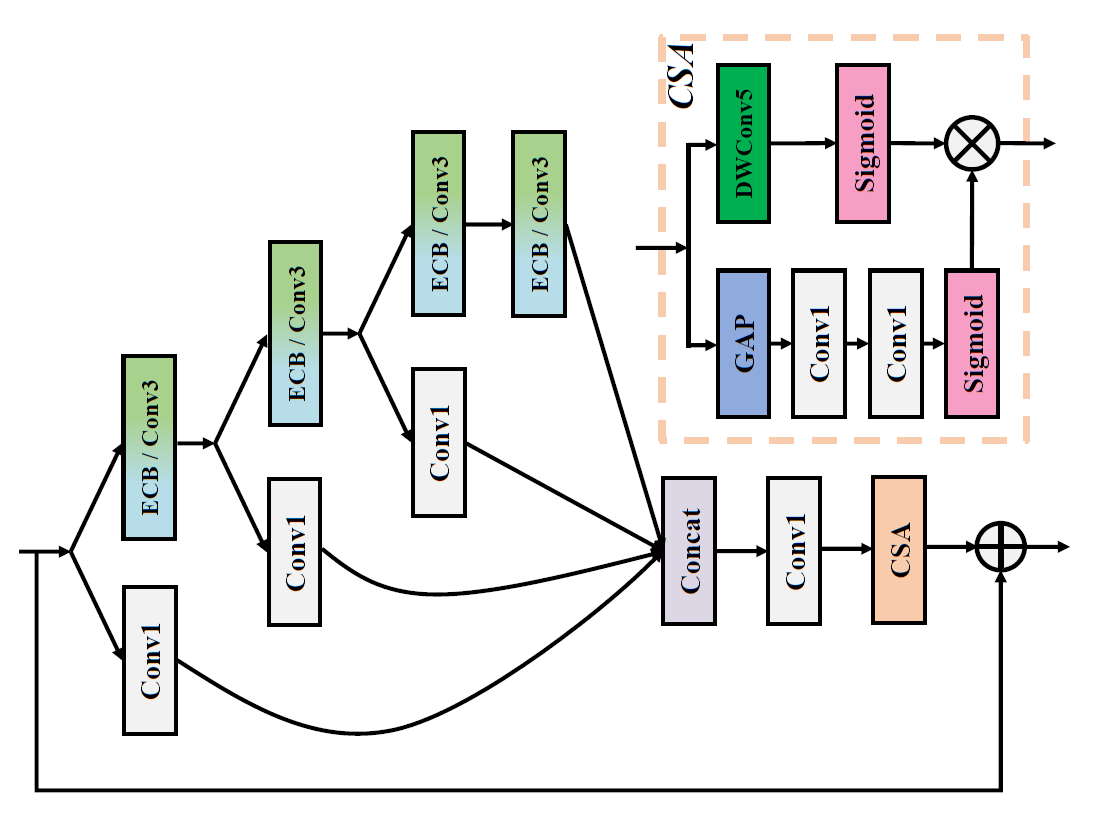}
\includegraphics[width=0.9\linewidth]{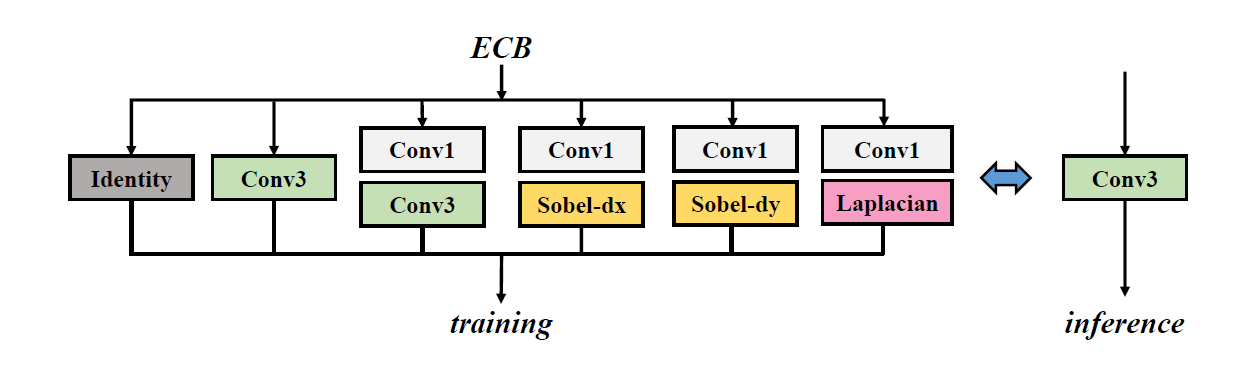}
\caption{\small{The RFD-CSA architecture proposed by team HITZST01.}}
\label{fig:HITZST01}
\end{figure}

Team HITZST01 proposed the RFD-CSA architecture~\cite{wu2022residual} for the considered problem demonstrated in Fig.~\ref{fig:HITZST01}. The model consists of there modules: the Source Features Module, the Enhance Features Module and the Upsample Features Module. The purpose of the Source Features Module is to extract rough features from the original raw images. This module is based on the ECB/Conv3 architecture proposed in~\cite{zhang2021edge} with the re-parameterization technique used to boost the performance while keeping the architecture efficient.

The Enhance Features Module is designed to extract effective features at multiple model levels. This module consists of $n$ lightweight multi-level feature extraction structures (with $n$ equal to 2 and 3 for models submitted to the 1st and 2nd challenge tracks, respectively). These structures are modified Residual feature distillation blocks~(RFDB) proposed in~\cite{liu2020residual}, where the CCA-layer in RFDB is replaced with the CSA block. The authors additionally added a long-term residual connection in this module to avoid the performance degradation caused by model's depth and to boost its efficiency. The Upsample Features Module generates the final image reconstruction results. To improve the performance, ECB/Conv3 blocks are added at the beginning and at the end of this module. The model was trained with a combination of the Charbonnier and SSIM losses using the Adam optimizer with the initial learning rate set to 1e--4 and halved every 200 epochs.

\subsection{MINCHO}

\begin{figure}[h!]
\centering
\resizebox{1.0\linewidth}{!}
{
\includegraphics[width=1.0\linewidth]{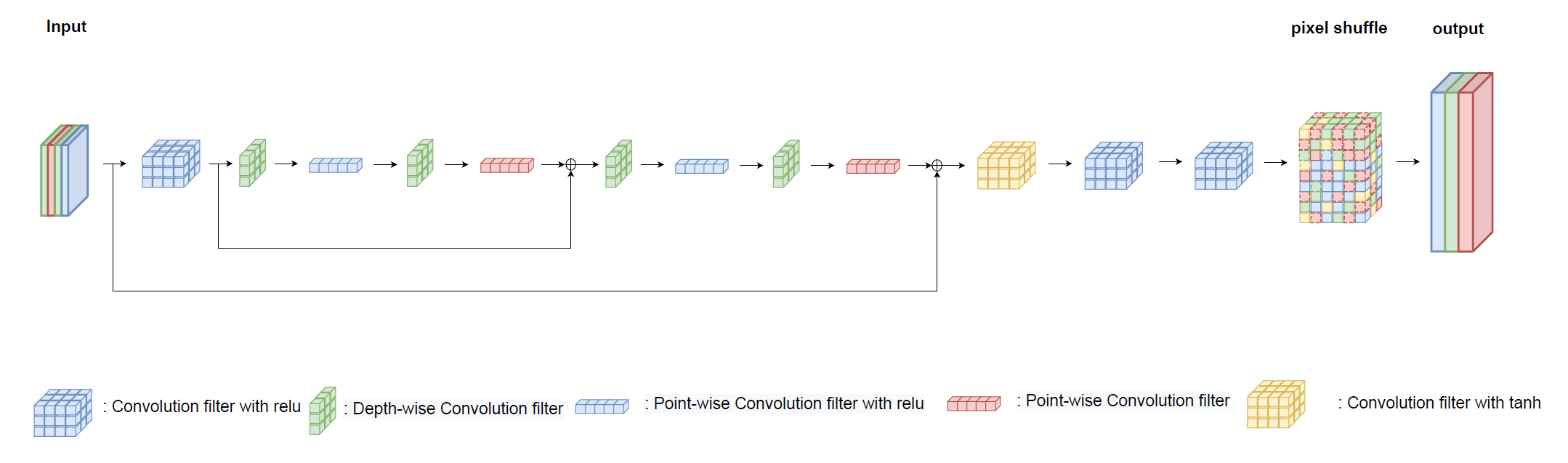}
}
\caption{\small{The architecture proposed by team MINCHO.}}
\label{fig:MINCHO}
\end{figure}

The architecture of the model developed by team MINCHO consists of the two main parts (Fig.~\ref{fig:MINCHO}). The first part is used to extract image features and consists of one convolutional blocks with \textit{ReLU} activations, depthwise convolutional blocks, pointwise convolutional blocks with \textit{ReLU} activations, and a skip connection. The second part of the model is the ISP part that is based on the Smallnet architecture~\cite{ignatov2021learned} with three convolutional and one pixel-shuffle layer. The model was first trained with L1 loss only, and then fine-tuned with a combination of the L2, perceptual-based VGG-19 and SSIM loss functions. Model parameters were optimized using the ADAM algorithm with $\beta_1=0.9, \beta_2=0.99, \epsilon=10^{-8}$, a learning rate of $10^{-4}$, and a batch size of 32.

\subsection{CASIA 1st}

\begin{figure}[h!]
\centering
\includegraphics[width=0.8\linewidth]{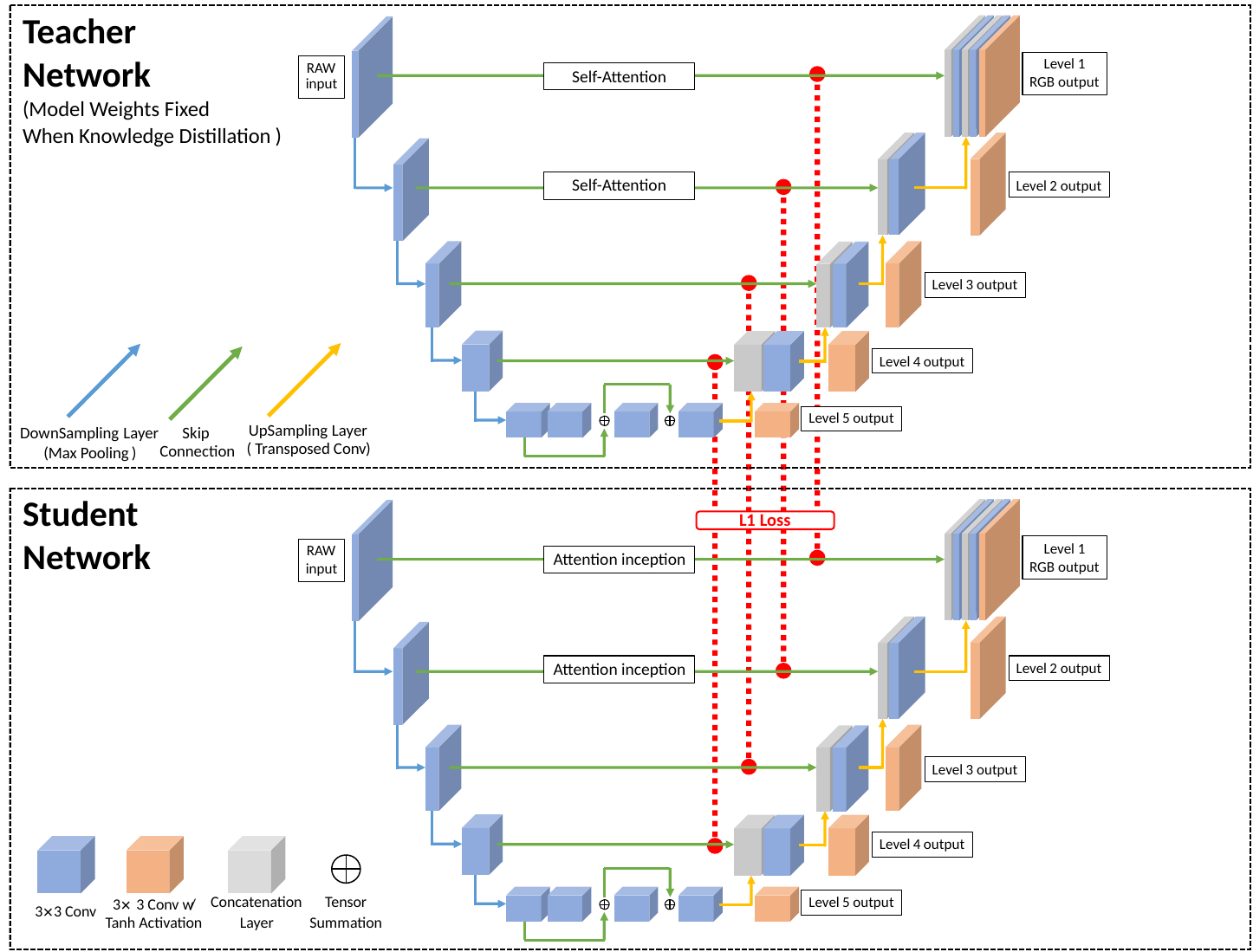}
\caption{\small{An overview of the model and teacher-guided training strategy proposed by team CASIA 1st.}}
\label{fig:CASIA}
\end{figure}

Team CASIA 1st proposed a two-stage teacher-guided model training strategy for the learned ISP problem (Fig.~\ref{fig:CASIA}). At the first stage, the authors trained a teacher network (TN) to get good PSNR and SSIM scores. They used the PUNET network~\cite{ignatov2021learned} as a baseline to develop an attention-aware based Unet (AA-Unet) architecture utilizing a self-attention module. Inspired by~\cite{ignatov2020replacing}, the authors applied a multi-loss 
constraint to features at different model levels. 

In the second stage, the authors designed a relatively tiny student network (SN) to inherit knowledge from the teacher network through model distillation. The student network is very similar to the teacher model. Inspired by~\cite{szegedy2017inception}, the authors connected three attention modules (CBAM, ECA, and SEA) in parallel to form the attention inception module that was used to replace all self-attention modules in the TN to reduce the time complexity. Since the self-attention and attention inception modules have the same input/output shapes, the model distillation strategy was straightforward: the feature maps from each level in the TN model were extracted and used as soft-labels to train SN. 

The models were trained with the MSE, L1, SSIM, VGG and edge~\cite{seif2018edge} loss functions. When training the TN, they applied MSE loss to level 2, 3 and 4, SSIM loss to level 2 and 3, and MSE, SSIM, VGG, Edge loss to level 1. When training SN, they fixed the model weight of TN and applied L1 loss to each level with the soft-label. Similar to TN, MSE, SSIM and VGG loss are also applied to level 1. Both TN and SN model parameters were optimized using Adam with a batch size of 12, a learning rate of 5e--5 and a weight decay of 5e--5.

\subsection{JMU-CVLab}

\begin{figure}[h!]
\centering
\includegraphics[width=0.75\linewidth]{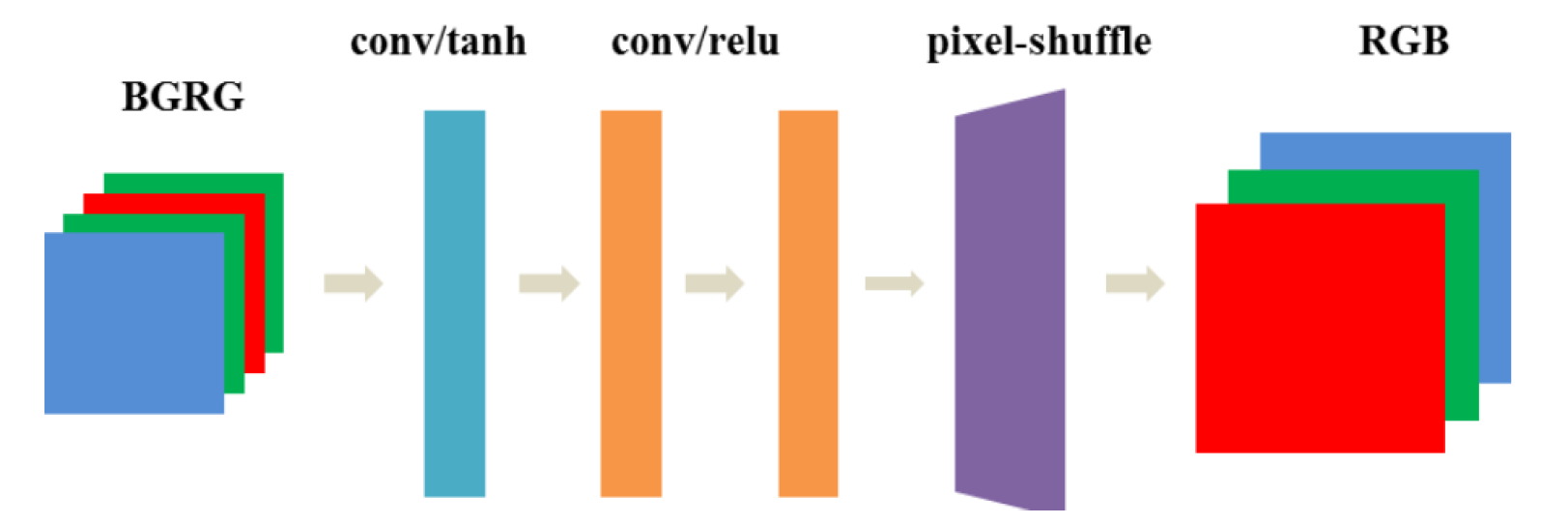}
\caption{\small{Model architecture used by the JMU-CVLab team.}}
\label{fig:JMU}
\end{figure}

Team JMU-CVLab based their solution on the Smallnet model developed in the previous MAI 2021 challenge~\cite{ignatov2021learned} (Fig.~\ref{fig:JMU}).
The authors replaced the PixelShuffle layer with the Depth2Space op and added two well-known non-linear ISP operations: tone mapping
and gamma correction~\cite{conde2022model}. These operations were applied to the final RGB reconstruction result followed by an additional CBAM attention block~\cite{woo2018cbam} applied to allow the model to learn more complex features. The model was trained to minimize the L1 and SSIM losses for 60 epochs with the Adam optimizer. A batch size of 32 with a learning rate of 0.0001 was used, basic augmentations including flips and rotations were applied to the training data.

\subsection{DANN-ISP}

\begin{figure}[h!]
\centering
\resizebox{1.0\linewidth}{!}
{
\includegraphics[width=1.0\linewidth]{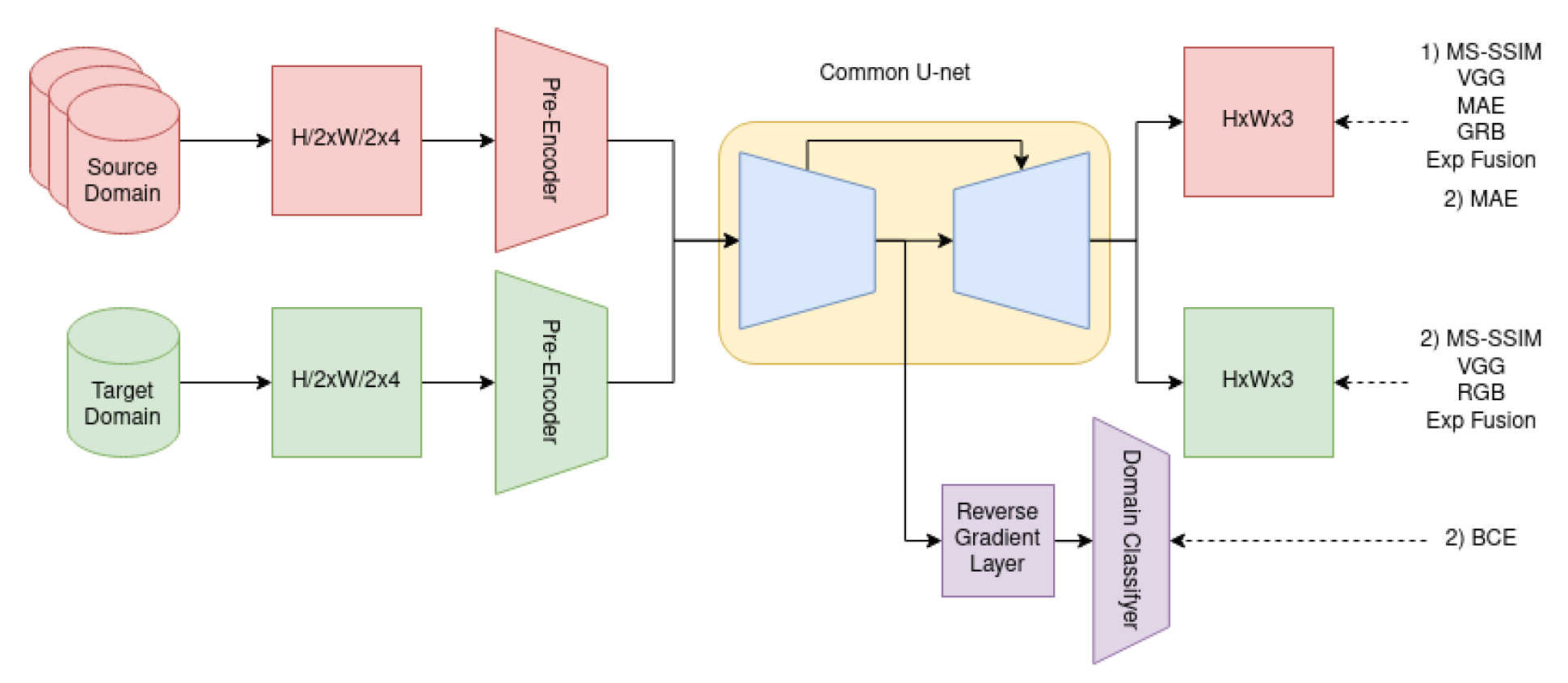}
}
\caption{\small{Domain adaptation strategy developed by team DANN-ISP.}}
\label{fig:DANN}
\end{figure}

Team DANN-ISP developed a domain adaptation method for the considered problem, which is illustrated in Fig.~\ref{fig:DANN}. The authors trained their model to generate RGB images with both source and target domains as inputs. Two pre-encoders were used to reduce the significant domain gap between different cameras by extracting individual and independent features from each one. The architecture of the pre-encoders consisted of three $3\times3$ convolutional layers with 8, 16 and 32 filters, respectively. Next, the authors used  a lightweight U-Net-like~\cite{ronneberger2015u} autoencoder with three downsampling and four upsampling blocks. It takes 32-channel features from each pre-encoder as input and produces two outputs: a 3-channel RGB image and a 256-dimensional feature vector from its bottleneck. Finally, a binary domain classifier~\cite{ganin2016domain} with an inverse gradient~\cite{ganin2015unsupervised} was used to reduce the gap between domains and increase the performance of the model. This classifier was constructed from the global average pooling layer and two dense layers at the end.

First, the model was pre-trained using only source domain data (in this case~-- using the Zurich-RAW-to-RGB dataset~\cite{ignatov2020replacing}) with the L1, MS-SSIM, VGG-based, color and exposure fusion losses. At the second stage, both source and target domain data were used together, and the model was trained to minimize a combination of the above losses plus the binary domain classifier loss (cross-entropy).

\subsection{Rainbow}

\begin{figure}[h!]
\centering
\includegraphics[width=0.75\linewidth]{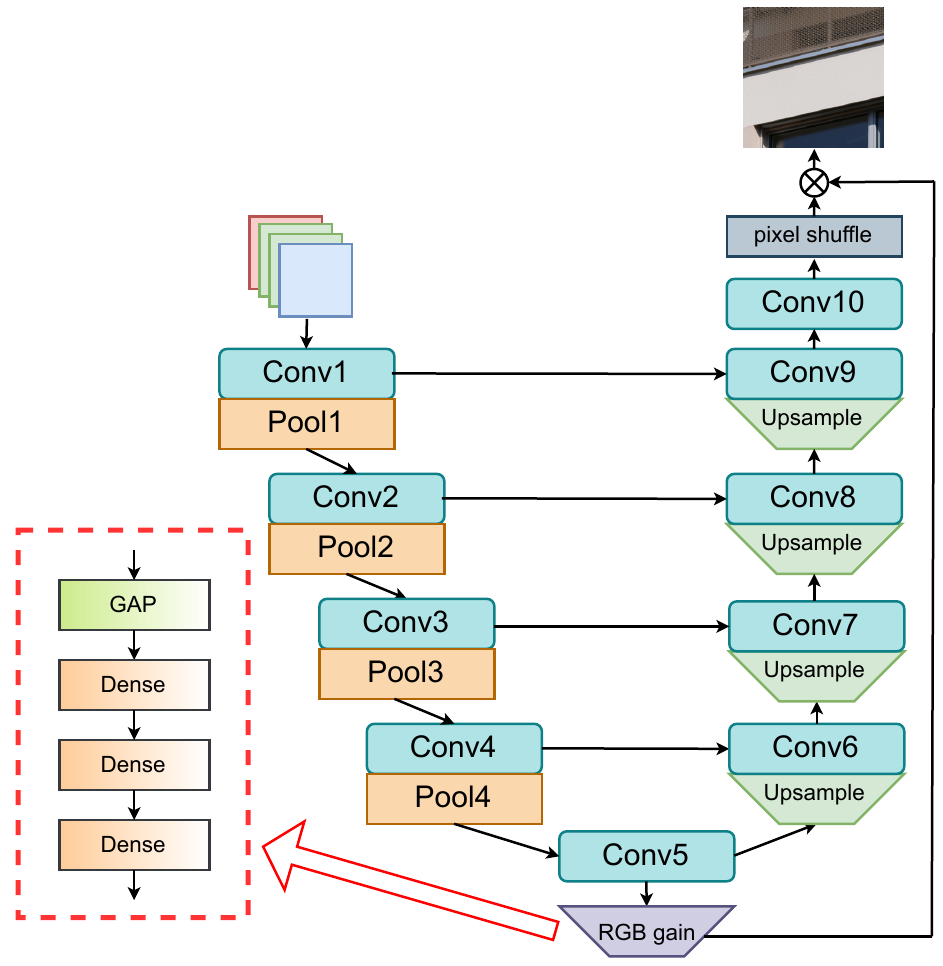}
\caption{\small{AWBUnet model proposed by team rainbow.}}
\label{fig:rainbow}
\end{figure}

Team rainbow proposed a U-Net based AWBUnet model for this task. In traditional ISP pipelines, demosaicing can be solved by a neural network naturally, and white balance needs to be determined based on the input image. Therefore, the authors propose an RGB gain module (Fig.~\ref{fig:rainbow}) consisting of a global average pooling (GAP) layer and three fully connected layers to adjust the white balance (RG channels) and brightness (RGB channels). The model was trained using a mini-batch size of 256, random horizontal flips were applied for data augmentation. The model was trained to minimizing the MSE loss function using the Adam optimizer. The initial learning rate was set to 2e--4 and halved every 100K iterations.

\subsection{SKD-VSP}

\begin{figure}[h!]
\centering
\resizebox{1.0\linewidth}{!}
{
\includegraphics[width=1.0\linewidth]{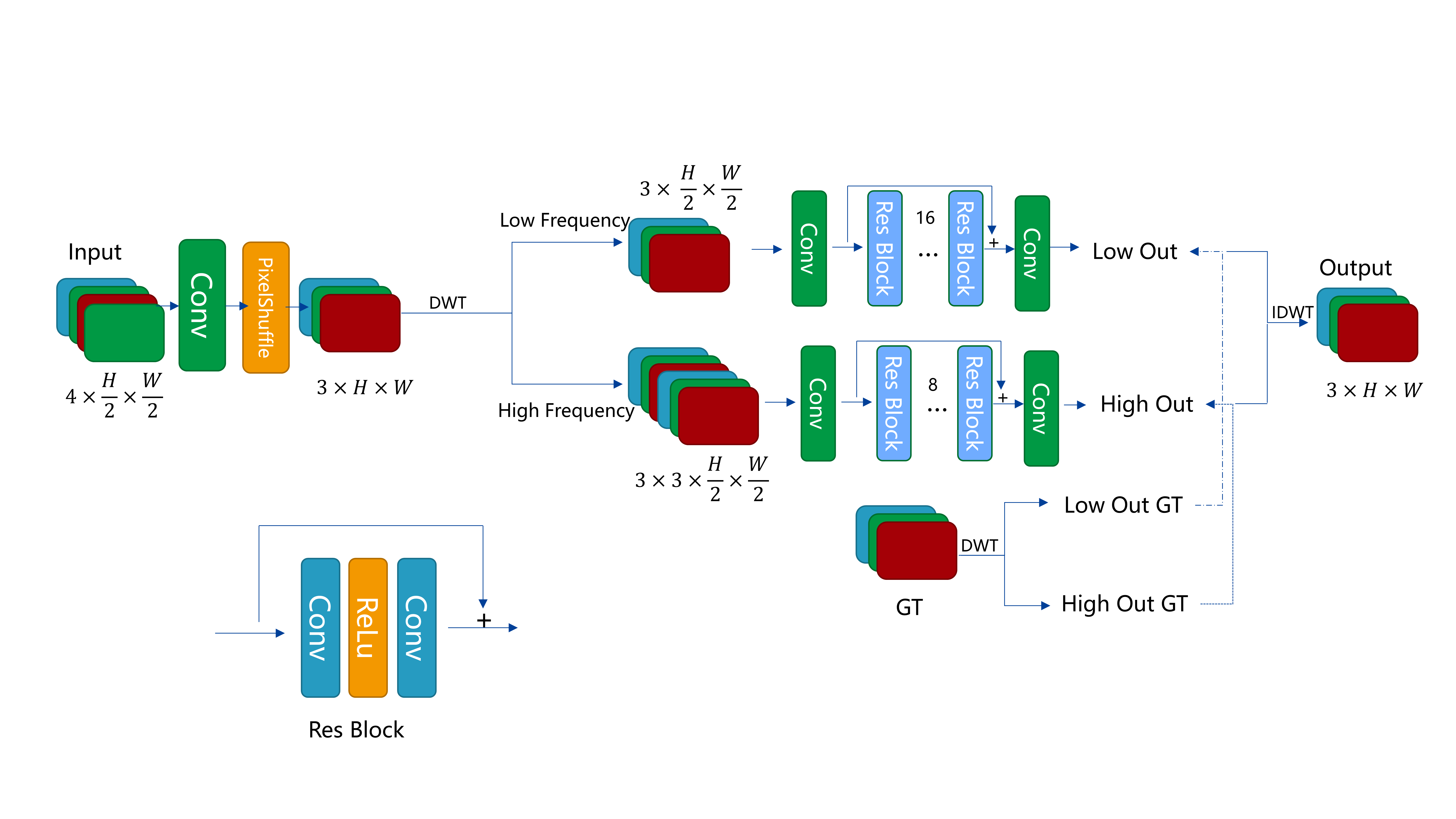}
}
\caption{\small{The architecture of the model proposed by team SKD-VSP. DWT stands for the discrete wavelet transform.}}
\label{fig:SKD}
\end{figure}

Team SKD-VSP applied the discrete wavelet transform to divide the image into high-frequency and low-frequency parts, and used two different networks to process these two parts separately (Fig.~\ref{fig:SKD}). After processing, the inverse discrete wavelet transform was used to restore the final image. The architecture of the considered networks is based on the EDSR~\cite{lim2017enhanced} model. Since the low-frequency part contains most of the global image information, the authors used 16 residual blocks with 256 channels in each block. For the high frequency part, 8 residual blocks with 64 channels were used due to its simple structure. When DWT operates on the image, the high-frequency part is divided into high-frequency information in three directions: horizontal, vertical and diagonal, so the input and output shape of the high-frequency network is $3\times 3\times \frac{H}{2}\times \frac{W}{2}$.

The authors used the L1 Loss for training the high frequency part, while for the low frequency part a combination of the L1, perceptual VGG-based and SSIM losses were utilized. Adam optimizer was used to train the model for 100 epochs with the initial learning rate set to 1e--3 and attenuated by a factor of 0.5 every 25 epochs.

\subsection{CHannel Team}

\begin{figure}[h!]
\centering
\resizebox{1.0\linewidth}{!}
{
\includegraphics[width=0.5\linewidth]{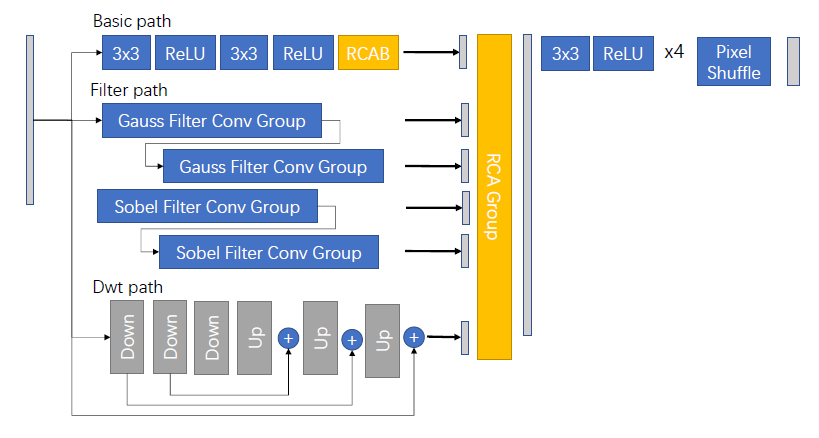}
\includegraphics[width=0.4\linewidth]{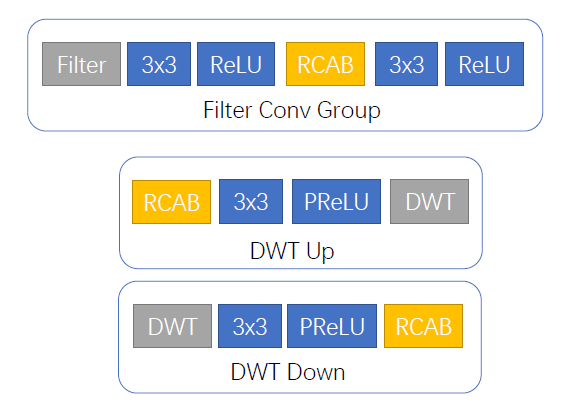}
}
\caption{\small{GaUss-DWT model architecture proposed by the CHannel Team.}}
\label{fig:CHannel}
\end{figure}

CHannel Team proposed a multi-path GaUss-DWT model, which architecture is illustrated in Fig.~\ref{fig:CHannel}. The model uses three paths to extract different information aspects from raw images: the basic path, the filter path, and the DWT path. The basic path consists of two convolutional layers and a residual attention block. This simple structure provides the model with a basic understanding of the image. The filter path consists of two different filters: Gauss and Sobel that extract chroma and texture information separately. Each convolutional group contains a fixed filter, two convolutional layers, and a residual channel attention block. The DWT path was inspired by the MW-ISPNet~\cite{ignatov2020aim} model that attains high fidelity results in the AIM 2020 challenge. However, its structure was simplified by replacing its residual channel attention groups with residual channel attention blocks. This modification allowed to reduce the runtime of the model while attaining satisfactory results. Finally, a residual channel attention group integrates the extracted information from all three paths. Its output is then processed by four convolutional layers and a pixel shuffle layer to get the final image result. The model was trained with a combination of the perceptual VGG-based, MSE, L1, MS-SSIM and edge loss functions. Model parameters were optimized using the Adam algorithm with a learning rate of 1e--4 and a batch size of 8.

\section{Additional Literature}

An overview of the past challenges on mobile-related tasks together with the proposed solutions can be found in the following papers:

\begin{itemize}
\item Learned End-to-End ISP:\, \cite{ignatov2019aim,ignatov2020aim,ignatov2022microisp,ignatov2022pynetv2}
\item Perceptual Image Enhancement:\, \cite{ignatov2018pirm,ignatov2019ntire}
\item Bokeh Effect Rendering:\, \cite{ignatov2019aimBokeh,ignatov2020aimBokeh}
\item Image Super-Resolution:\, \cite{ignatov2018pirm,lugmayr2020ntire,cai2019ntire,timofte2018ntire}
\end{itemize}

\section*{Acknowledgements}

We thank the sponsors of the Mobile AI and AIM 2022 workshops and challenges: AI Witchlabs, MediaTek, Huawei, Reality Labs, OPPO, Synaptics, Raspberry Pi, ETH Z\"urich (Computer Vision Lab) and University of W\"urzburg (Computer Vision Lab).

\appendix
\section{Teams and Affiliations}
\label{sec:apd:team}

\bigskip

\subsection*{Mobile AI 2022 Team}
\noindent\textit{\textbf{Title: }}\\ Mobile AI 2022 Learned Smartphone ISP Challenge\\
\noindent\textit{\textbf{Members:}}\\ Andrey Ignatov$^{1,2}$ \textit{(andrey@vision.ee.ethz.ch)}, Radu Timofte$^{1,2,3}$\\
\noindent\textit{\textbf{Affiliations: }}\\
$^1$ Computer Vision Lab, ETH Zurich, Switzerland\\
$^2$ AI Witchlabs, Switzerland\\
$^3$ University of Wuerzburg, Germany\\

\subsection*{MiAlgo}
\noindent\textit{\textbf{Title:}}\\ 3Convs and BigUNet for Smartphone ISP\\
\noindent\textit{\textbf{Members: }}\\ \textit{Shuai Liu (liushuai21@xiaomi.com)}, Chaoyu Feng, Furui Bai, Xiaotao Wang, Lei Lei\\
\noindent\textit{\textbf{Affiliations: }}\\
Xiaomi Inc., China\\

\subsection*{Multimedia}
\noindent\textit{\textbf{Title:}}\\FGARepNet: A real-time end-to-end ISP network based on Fine-Granularity attention and Re-parameter convolution\\
\noindent\textit{\textbf{Members:}}\\ \textit{Ziyao Yi (yi.ziyao@sanechips.com.cn)}, Yan Xiang, Zibin Liu, Shaoqing Li, Keming Shi, Dehui Kong, Ke Xv \\
\noindent\textit{\textbf{Affiliations: }}\\
Sanechips Co. Ltd, China\\

\subsection*{ENERZAi Research}
\noindent\textit{\textbf{Title:}}\\Latency-Aware NAS and Histogram Feature Loss\\
\noindent\textit{\textbf{Members:}}\\ \textit{Minsu Kwon (minsu.kwon@enerzai.com)}\\
\noindent\textit{\textbf{Affiliations: }}\\
ENERZAi, Seoul, Korea \\ \textit{enerzai.com}\\

\subsection*{HITZST01}
\noindent\textit{\textbf{Title:}}\\Residual Feature Distillation Channel Spatial Attention Network for ISP on Smartphones~\cite{wu2022residual}\\
\noindent\textit{\textbf{Members:}}\\ \textit{Yaqi Wu$^1$ (titimasta@163.com)}, Jiesi Zheng$^2$, Zhihao Fan$^3$, Xun Wu$^4$, Feng Zhang\\
\noindent\textit{\textbf{Affiliations: }}\\
$^1$ Harbin Institute of Technology, China\\
$^2$ Zhejiang University, China\\
$^3$ University of Shanghai for Science and Technology, China\\
$^4$ Tsinghua University, China\\

\subsection*{MINCHO}
\noindent\textit{\textbf{Title:}}\\Mobile-Smallnet: Smallnet with MobileNet blocks for an end-to-end ISP Pipeline\\
\noindent\textit{\textbf{Members:}}\\ \textit{Albert No (albertno@hongik.ac.kr)}, Minhyeok Cho\\
\noindent\textit{\textbf{Affiliations: }}\\
Hongik University, Korea\\

\subsection*{CASIA 1st}
\noindent\textit{\textbf{Title:}}\\Learned Smartphone ISP Based On Distillation Acceleration\\
\noindent\textit{\textbf{Members:}}\\ \textit{Zewen Chen$^1$ (chenzewen2022@ia.ac.cn)}, Xiaze Zhang$^2$, Ran Li$^3$, Juan Wang$^1$, Zhiming Wang$^4$\\
\noindent\textit{\textbf{Affiliations: }}\\
$^1$ Institute of Automation, Chinese Academy of Sciences, China\\
$^2$ School of Computer Science, Fudan University, China\\
$^3$ Washington University in St. Louis\\
$^4$ Tsinghua University, China\\

\subsection*{JMU-CVLab}
\noindent\textit{\textbf{Title:}}\\Shallow Non-linear CNNs as ISP\\
\noindent\textit{\textbf{Members:}}\\ \textit{Marcos V. Conde (marcos.conde-osorio@uni-wuerzburg.de)}, Ui-Jin Choi\\
\noindent\textit{\textbf{Affiliations: }}\\
University of Wuerzburg, Germany\\

\subsection*{DANN-ISP}
\noindent\textit{\textbf{Title:}}\\Learning End-to-End Deep Learning Based Image Signal Processing Pipeline Using Adversarial Domain Adaptation\\
\noindent\textit{\textbf{Members:}}\\ \textit{Georgy Perevozchikov (perevozchikov.gp@phystech.edu)}, Egor Ershov\\
\noindent\textit{\textbf{Affiliations: }}\\
Moscow Institute of Physics and Technology, Russia\\

\subsection*{Rainbow}
\noindent\textit{\textbf{Title:}}\\Auto White Balance UNet for Learned Smartphone ISP\\
\noindent\textit{\textbf{Members:}}\\ \textit{Zheng Hui (huizheng.hz@alibaba-inc.com)}\\
\noindent\textit{\textbf{Affiliations: }}\\
Alibaba DAMO Academy, China\\

\subsection*{SKD-VSP}
\noindent\textit{\textbf{Title:}}\\IFS Net-Image Frequency Separation Residual Network\\
\noindent\textit{\textbf{Members:}}\\ \textit{Mengchuan Dong (mengchuan61@gmail.com)}, Wei Zhou, Cong Pang\\
\noindent\textit{\textbf{Affiliations: }}\\
ShanghaiTech University, China\\

\subsection*{CHannel Team}
\noindent\textit{\textbf{Title:}}\\GaUss-DWT net\\
\noindent\textit{\textbf{Members:}}\\ \textit{Haina Qin (qinhaina2020@ia.ac.cn)}, Mingxuan Cai\\
\noindent\textit{\textbf{Affiliations: }}\\
Institute of Automation, Chinese Academy of Sciences, China\\

{\small
\bibliographystyle{splncs04}
\bibliography{egbib}
}

\end{document}